\documentclass[twocolumn]{svjour3}
\pdfoutput=1
\usepackage{times} 
\usepackage[utf8]{inputenc}
\usepackage{hyphenat} 
\usepackage{graphicx}
\graphicspath{{./plots/}}
\usepackage{afterpage}
\usepackage{amsmath} 
\usepackage{amssymb} 
\usepackage{mathtools}
\usepackage{bigints}
\newcommand{\breakingcomma}{%
  \begingroup\lccode`~=`,
  \lowercase{\endgroup\expandafter\def\expandafter~\expandafter{~\penalty0 }}}

\usepackage{breqn}
\usepackage{microtype}
\usepackage{bm}
\usepackage{algorithm}
\usepackage{algpseudocode}
\usepackage{graphicx}
\usepackage{stfloats}

\usepackage{tikz}
\usepackage{xcolor}
\usepackage{pgfplots}
\pgfplotsset{compat=1.14}
\usepgfplotslibrary{groupplots}
\usepackage{booktabs}

\usepgfplotslibrary{external} 
\usetikzlibrary{external} 
\makeatletter
\let\c@author\relax
\makeatother

\usepackage[sorting=none,natbib=true]{biblatex}

\begin{document}
\title{MOrdReD: Memory-based Ordinal Regression Deep Neural Networks for Time Series Forecasting}

\author{Bernardo P\'erez Orozco \and Gabriele Abbati \and Stephen J. Roberts}
\institute{\textbf{Corresponding author:} \textit{B. P\'erez Orozco} 
\at Information Engineering, Dept. of Engineering Science, University of Oxford, UK\\
\textit{Tel:} +44 (0)1865 616600\\
\textit{ORCID:} 0000-0001-9741-7950\\
\email{ber@robots.ox.ac.uk}
\\ 
\textit{G. Abbati} \at Information Engineering, Dept. of Engineering Science, University of Oxford,  UK\\\email{gabb@robots.ox.ac.uk}
\\
\textit{S.J. Roberts} \at Information Engineering, Dept. of Engineering Science \& Oxford-Man Institute of Quantitative Finance, University of Oxford, UK\\
\& Mind Foundry Ltd.\\\email{sjrob@robots.ox.ac.uk}}

\date{Received: 24 October 2018 / Accepted: TBD}
\maketitle

\begin{abstract}
Time series forecasting is ubiquitous in the modern world. Applications range from health care, astronomy and include climate modelling, to financial trading and monitoring of critical engineering equipment. To offer robust value over this wide range of activities, models must not only provide accurate forecasts, but also quantify and adjust their uncertainty over time. In this work, we directly tackle both tasks with a novel, fully end-to-end deep learning method. By recasting time series prediction as an ordinal regression task, we develop a principled methodology to assess long-term predictive uncertainty and describe the rich, multi-modal, non-Gaussian behaviour which arises regularly in many problem domains. 

Notably, our framework is a wholly general-purpose approach that requires little to no user intervention to be used. We showcase this key feature in a large-scale benchmark test with 45 datasets drawn from both a wide range of real-world applications and a comprehensive list of synthetic data. This wide comparison uses, as benchmark methods, state-of-the-art models from both Machine Learning and Statistics literature, such as the Gaussian Process. We find that our approach not only provides excellent predictive forecasts and associated uncertainty bounds, but also allows us to infer derived information, such as the predictive distribution of critical events of interest. We show that the method we propose performs accurately and reliably, even over long time horizons. 
 
\end{abstract}
    
	\section{INTRODUCTION}
\label{sec:intro}

Performing long-term, reliable time series forecasting has wide interest in both academia and industry. Application domains are great in number, including meteorology, energy systems, astronomy, finance, dynamical systems, physiology and many others. Crucially, forecasts about a system's future should not be restricted to merely investigating the most likely, or expected, values of the system at a specific time instant. Information regarding the likely timing of an event of interest is important, and so are beliefs regarding the value at a given time. Such questions can naturally be answered by providing full predictive probability distributions at each time step. 


In the main, forming a rich probability distribution over future states has been addressed by statistical models. The latter are often parametric, such as polynomial regressors or the family of Markovian state-space models. More recently, non-parametric approaches such as Gaussian Processes (GPs) \citep{Rasmussen2006} allow avoidance of pre-specified parametric forms. In part, no doubt, due to their native ability to infer full predictive distributions, GPs (and related models) have increasingly dominated the time series modelling literature over the last decade, especially as  classic Markov models (including autoregressive processes and the like) can be seen as special cases of the GP. We simply note that, powerful though the GP approach is, the standard GP model makes the tacit assumption that the residuals are normally distributed and hence the conditional predictive posterior is also Gaussian. This could be restrictive in the long-term forecasting scenario, where short-term forecast uncertainty may induce distant scenarios whose nature is inherently multi-modal. Furthermore, in its naive form, the GP also suffers from poor scalability, which can hinder the use of the large amounts of data readily available in some tasks. Lastly, to obtain best performance, expert knowledge, or extensive searches may be needed to refine the kernel function at the heart of the GP.

In parallel to these developments, deep and recurrent neural network models have gained significant attention in the domain of classification tasks, such as those that arise in the Natural Language Processing community. Neural networks, despite their complexity, scale well with dataset size and the increasing availability and development of algorithms, data, hardware and software have enabled these models to become the state-of-the-art in sequential classification tasks such as speech recognition, machine translation, scene labelling, and others. In the context of regression tasks such as time series forecasting, however, these models have not achieved the same success. Even though time series forecasting with neural networks has been investigated extensively in the literature, to the authors' best knowledge, this work only deals with point forecasts. A potential reason is that fully probabilistic models such as Bayesian neural networks are analytically intractable, and approximate Bayesian inference methods can be computationally expensive even for networks with relatively few units. This leaves little room to justify their usage in comparison with probabilistic methods.

In this work, we develop a recurrent neural network framework for time series prediction that is capable of producing fully (though approximate) probabilistic forecasts. We achieve this by recasting the time series forecasting task as an ordinal (auto-)regression one, which consists of passing the time series through a quantisation step, thus transforming the prediction task into a sequential classification problem. This enables us to incorporate state-of-the-art developments from the Machine Learning and Natural Language Processing communities. Such ordinal forecasts can further describe rich multi-modal, non-Gaussian behaviour, akin to that which may describe long-term forecasts more faithfully. The number of modes and other shape features of the predictive posterior distribution can thence adapt dynamically for different predictive horizons.

Moreover, our ordinal framework is a wholly general-purpose approach that is scalable with dataset size and requires little to no human intervention or expert craftsmanship for model specification. We showcase this key feature by means of an exhaustive comparison over 45 datasets from a wide range application domains, encompassing a variety of state-of-the-art methods from both the Machine Learning and Statistical modelling communities. Crucially, we evaluate both the accuracy and calibration of our probabilistic forecasts, in addition to the deviation of point forecasts, such as the mean and the median of the predictive distributions. We provide evidence that our method achieves superior performance in the majority of datasets, and furthermore, we show that even when it does not, on average we still expect it to surpass most of its competitors.

To further showcase how our probabilistic forecasts can be used to address other queries of interest, we also evaluate our model in the context of event timing prediction. In particular, we demonstrate how to construct predictive densities for the occurrence of local maxima with Kernel Density Estimation, and show that our framework is able to accurately form beliefs over the timing of such events.

In summary, the contributions of our work are as follows:
\begin{itemize}
	\item We develop a fully end-to-end framework for long-term time series forecasting that requires little to no manual model tuning and outperforms other state-of-the-art techniques;
	\item we incorporate a sequential procedure to infer reliable long-term uncertainty bounds in an autoregressive fashion;
	\item we perform an exhaustive, large-scale comparison study of this and other state-of-the-art forecasting techniques;
	\item we demonstrate how our probabilistic forecasts can be employed to predict the occurrence of (e.g.) local maxima in a timely and reliable manner.
\end{itemize}

The remainder of this paper is organised as follows: in Section \ref{sec:lit} we offer a review of the relevant literature; then, in Section \ref{sec:methods} we present an overview of our ordinal autoregression framework, as well as a succinct review of the methods we used as benchmarks; in Section \ref{sec:experiments} the experiments that support the paper and the data used are detailed. We lastly present our conclusions in Section \ref{sec:conclusions}.

	\section{LITERATURE REVIEW} \label{sec:lit}

\subsection{MACHINE LEARNING METHODS FOR TIME SERIES FORECASTING} \label{subsec:mltsf}

Machine Learning (ML) has a rich history in the time series literature, covering a broad range of methods such as random forest regressors \citep{breiman2001random}, quantile random forests \citep{meinshausen2006quantile} and artificial neural networks (ANNs) \citep{Bishop1995}. However, many of these do not readily provide probabilistic forecasts, and some others only do so in a limited fashion. 

Consequently, the time series literature has increasingly been dominated by models such as the Gaussian Process \cite{Hu2015,Wu2012,Laib2018,Roberts2013,Kong2018} which enable the inference of full predictive distributions in a principled manner, even in the presence of little data. Indeed, even summarised predictions, such as those given by the predictive posterior mean, have been shown to produce smaller deviations than those provided by competitors such as neural networks in a variety of application domains, for example in \cite{Pasolli2010,Liu2018,Kamath2018}. Their standard form, however, assumes that residuals distribute normally and thence the conditional predictive posterior is also a Gaussian. Furthermore, this approach (in its most general form) suffers from poor scalability, which can hinder the use of large datasets.

Despite not being able to produce probabilistic forecasts by design, ANNs are popular ML models that are known to scale well with dataset size. As a result, there is a rich literature covering applications of time series forecasting with feed-forward neural network models such as the multi-layer perceptron (MLP) and the radial basis function network (RBFNs), for example: electricity pricing \cite{Bisoi2018}, stock pricing \cite{Zhang2018}, water quality \cite{Faruk2010} and solar cell energy output \cite{Awad2018}. Some neural network architectures are also endowed with a memory mechanism, as is the case for the recurrent neural networks (RNNs). This mechanism refers to an encoding scheme in which the correlation between successive elements in sequential data is exploited to extract characteristic temporal features. RNNs however suffer from the vanishing and exploding gradient problems, which renders them more difficult to optimise by gradient descent than their memory-less counterparts. This has been addressed by introducing learnable gating mechanisms that allow for error signals to be propagated through time, thus enabling efficient learning via gradient descent, as in the case of the long short-term memory (LSTM) \citep{Hochreiter1997}. In consequence, LSTMs have shown themselves to offer excellent, end-to-end performance in predictions over symbol sequences, especially in the Natural Language Processing literature \citep{Graves2013,Sutskever2014,Xu2014}.

Gating mechanisms, however, come at the expense of introducing more model parameters, which in turn may require more data to be inferred without overfitting. Indeed, such networks could only be trained at small scale until recently, and this could be a potential reason why early findings showed that they were easily outperformed by simpler, memory-less methods \citep{Gers2001} in the context of time series. With the development of specialised hardware and software however, as well as increased availability of data for some tasks, recurrent models with a greater numbers of units have now been successfully used in several application domains, such as traffic flow prediction \cite{Ma2015,Fu2016}, precipitation nowcasting \cite{Shi2015}, chaotic dynamical systems \cite{Chandra2012,Ardalani2010}, electricity market price \cite{Anbazhagan2013}, financial market predictions \citep{Rutkauskas2011} and energy output forecasting \citep{Monteiro2013}.

\subsection{COMPUTING PREDICTIVE UNCERTAINTY WITH RECURRENT NEURAL NETWORKS} \label{subsec:nnunc}

In spite of its extent, the time series literature addressing the quantification and evaluation of predictive probability distributions with (recurrent or otherwise) neural networks remains scarce, with much being historic research from decades ago. Further, assessments are often constrained to benchmarking average deviation metrics, such as the mean squared or absolute errors (MSE, MAE) and the like. Some portion of the literature has attempted to quantify uncertainty, either in a limited fashion (for instance, by estimating upper and lower bounds \cite{ak2016}), or fully, by performing approximate or sample-based Bayesian inference to directly learn a Bayesian Neural Network \cite{Neal1996}. The latter approach however introduces scalability issues that make it difficult to justify their usage for time series prediction in lieu of models such as the Gaussian Process.

Nonetheless, recent findings linking the dropout regularisation technique to approximate Bayesian inference \citep{pmlr-v48-gal16,gal2016theoretically} have enabled an efficient and principled, though still approximate, quantification of uncertainty in a variety of settings. Indeed, recent work looks to apply these approaches to time series forecasting \citep{zhu2017deep}, however still making the tacit assumption that the predictive distribution is uni-modal Gaussian.

Dropout \citep{srivastava2014dropout} is a stochastic regularisation technique that has been successfully employed to prevent overfitting and approximately combine different network architectures. At training time, before each feed-forward operation (necessary to compute the gradients of the loss function), each unit of the neural network is dropped with probability $p_{\text{drop}}$. In this way, a mask computed accordingly to this Bernoulli distribution is placed on the network, which in turn gets ``thinned''. For gradient-based optimisers, gradients are computed with these new approximate outputs before updating the learnable parameters. A new mask is then drawn for each subsequent iteration. 

With this approach, a neural network with $n$ units can be seen as a collection $2^n$ thinned networks sharing a substantial number of weights. As shown in \citep{pmlr-v48-gal16}, a Monte Carlo approach can then be followed to approximate the predictive posterior distribution. This derives from the fact that, under certain assumptions, a neural network of arbitrary depth trained with dropout before each layer is mathematically equivalent to a probabilistic Deep Gaussian Process \citep{damianou2013deep}.



	\section{METHODOLOGY}
\label{sec:methods}

In this Section, we develop  the methods  we use for our large-scale benchmarking task. We start by introducing our own MOrdReD\footnote{In Arthurian legend, Mordred was the illegitimate son of King Arthur. Mordred fell in battle against his own father.} methodology: a Memory-endowed Ordinal Regression Deep neural network for time series forecasting. We then give a succinct summary of the baselines used to compare our model with, namely Gaussian Processes and the state-space formulation of autoregressive (AR) forecasting. 

\subsection{ORDINAL REGRESSION}\label{sub:ordred}

We start by defining what an ordinal regression task is. Consider a bounded time series with range $I \subset \mathbb{R}$ and let $C = \{C_i\}_{i=1}^{M}$ be a partition of $I$ with cardinality $M$. Without loss of generality, we assume the $C_i$ have all the same measure on $\mathbb{R}$ (e.g. the $C_i$ are non-overlapping sub-intervals of $I$ with equal sizes). 

We can then define the time series forecasting problem in an ordinal and autoregressive fashion. Assume the $P$-sample-long observed sequence $\mathbf{X}^{(t)}\in \mathbb{R}^{P \times M}$, with $\mathbf{X}^{(t)} = (\mathbf{x}_{1}^{(t)}, \dots, \mathbf{x}_{P}^{(t)}) = (\mathbf{x}_{t - P + 1}, \dots, \mathbf{x}_{t})$, where $P$ is the lookback window horizon and each $\mathbf{x}_i\in\mathbb{R}^M$ is the quantised one-hot encoded representation of a time series observation, originally assumed to be real-valued. The task of ordinal regression, which lies between regression and classification, consists in learning a map from the latter sequence to a symbol $\mathbf{y}^{(t)}$. Autoregressive time series forecasting can then be enabled by letting $\mathbf{y}^{(t)} = \mathbf{x}^{(t)}_{P+1}=\mathbf{x}_{t+1}$.

The procedure described above is indeed equivalent to framing the quantised one-hot encoded time series $\mathbf{x}$  with a sliding window of length $P$. Given the symbol set cardinality $M$ and lookback $P$, our time series datasets can be now defined as $\bm{\mathcal{X}} = \{\mathbf{X}^{(P)}, \dots, \mathbf{X}^{(N)}\}$ and $\bm{\mathcal{Y}} = \{\mathbf{y}^{(P)}, \dots, \mathbf{y}^{(N)}\}$.

We remark that such autoregression may induce a potential loss of accuracy during the quantisation phase. This can be mitigated by choosing a relatively large $M$ that allows for a sufficiently fine-grained partition $C$, and we note that, within reason, the choice of $M$ (problem dependent and a trade-off between resolution and computational simplicity) does not affect results. LSTMs (introduced in the next Section) thence naturally arise as a befitting methodology, due to their empirically proven ability to model categorical sequences over large symbol sets, such as those that frequently arise in the Natural Language Processing domain \citep{Sutskever2014}. We further note that, within reason, quantisation has little impact in other queries of interest, such as long-term trend changes or event occurrence forecasting.

\subsection{MEMORY-ENDOWED ORDINAL REGRESSION DEEP NEURAL NETWORKS}\label{sub:forecasting_lstm}

In this Section our forecasting method is introduced. We consider a Neural Network parameterised by a vector of parameters $\bm{\theta}$, whose output is $f^{\bm{\theta}}(\mathbf{X})$. In particular, we consider LSTM-based architectures trained on a dataset $\bm{\mathcal{X}}$, $\bm{\mathcal{Y}}$, with model likelihood $p(\mathbf{y} \mid \mathbf{X}, \bm{\theta})$.

\subsubsection{The Long Short Term Memory}\label{subsub:lstm}

The Long Short-Term Memory (LSTM) \citep{Hochreiter1997} is a particular type of recurrent neural network (RNN) endowed with a gating mechanism that enables efficient propagation of error signals in time. Such mechanisms prevent well-known gradient instability issues that arise in other simple recurrent neural network models \citep{Pascanu2012}. Gating thus allows for efficient gradient-based learning of temporal features, which are encoded internally in memory cells that are updated with every new observation in the sequence. 

Consider an observed sequence $\mathbf{X}^{(n)} = (\mathbf{x}_{1}^{(n)}, \dots, \mathbf{x}_{P}^{(n)})$, and let $\odot$ be the element-wise product operator. For each $t = 1,\dots,P$, the LSTM outputs a vector of temporal features $\mathbf{h}_t$, which is given by:
\begin{align*}
	\mathbf{h}_t &= \mathbf{o}_t \odot \tanh(\mathbf{C}_t).
\end{align*}
$\mathbf{C}_t$ is the memory cell at time $t$:
\begin{align*}
	\mathbf{C}_t = \mathbf{i}_t\odot\mathbf{S}_t + \mathbf{f}_t\odot\mathbf{C}_{t-1},
\end{align*}
where
\begin{align*}
	\mathbf{S}_t = \tanh{(\mathbf{W}_{\mathbf{S}} \mathbf{x}'_t + \mathbf{b}_{\mathbf{S}})}
\end{align*}
and $\mathbf{i}_t, \mathbf{o}_t, \mathbf{f}_t$ are the input, output and forget gates respectively:
\begin{align*}
	\mathbf{i}_t =& \sigma(\mathbf{W}_{\mathbf{i}} \mathbf{x}'_t + \mathbf{b}_{\mathbf{i}}),
	\\
	\mathbf{o}_t =& \sigma(\mathbf{W}_{\mathbf{o}} \mathbf{x}'_t + \mathbf{b}_{\mathbf{o}}),
	\\
	\mathbf{f}_t =& \sigma(\mathbf{W}_{\mathbf{f}} \mathbf{x}'_t + \mathbf{b}_{\mathbf{f}}).
\end{align*}
Here $\sigma$ is the element-wise sigmoid function and $\mathbf{x}'_t = (\mathbf{x}_t, \mathbf{h}_{t-1})$ is the concatenation of the observation $\mathbf{x}_t$ and the LSTM output at the previous time step $\mathbf{h}_{t-1}$. In the ordinal regression setting, we optimise the categorical cross-entropy loss function with respect to the LSTM's learnable parameters  $\bm{\theta} = \{\mathbf{W}_{\mathbf{i}}, \mathbf{W}_{\mathbf{o}}, \mathbf{W}_{\mathbf{f}}, \mathbf{W}_{\mathbf{S}}, \mathbf{b}_{\mathbf{i}}, \mathbf{b}_{\mathbf{o}}, \mathbf{b}_{\mathbf{f}}, \mathbf{b}_{\mathbf{S}}\}$.

\subsubsection{Sequence-to-Sequence}\label{subsub:seq2seq}

LSTMs can be further enriched by incorporating learnt features from the backwards temporal dynamics of the time series, in a bidirectional fashion. This is simply given by a second LSTM that scans the input sequence $\mathbf{X}$ in reverse order, and whose outputs are then averaged with those of the initial LSTM. This bidirectional model has shown remarkable success in the NLP literature \citep{graves2005}, including extensions to LSTM-based models such as the sequence-to-sequence architecture \citep{Sutskever2014}. 

\begin{figure*}
	\centering
	\includegraphics[width=\textwidth]{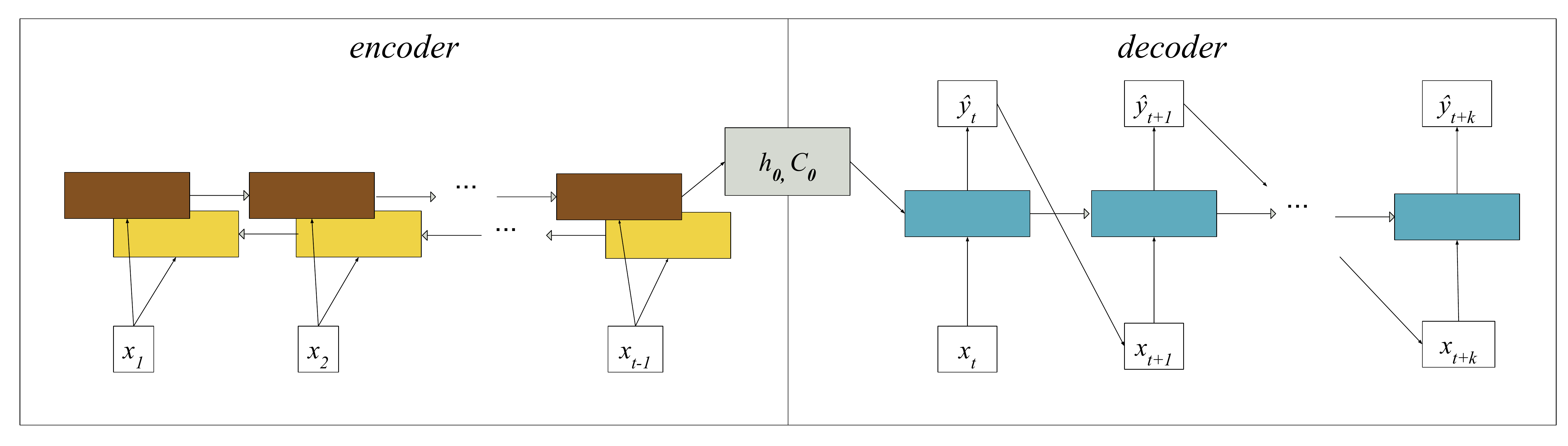}
	\caption{Visual depiction of a sequence-to-sequence model with a bidirectional encoder. The input sequence is scanned up to time index $t-1$ and subsequently summarised as the decoder's initial state $h_0$. The decoder then produces a one-step-ahead forecast for every new incoming input up to time $t+k$. At prediction time, autoregression is enabled by allowing $x_{t+i} = \hat{y}_{t+i-1}$.} 
	\label{fig:seq2seq}
\end{figure*}

We now focus our attention to the latter. The sequence-to-sequence model consists of two recurrent neural network models: an encoder $f^{(\text{enc})}$, which maps the observed sequence, $\mathbf{X}$, into a fixed-dimensional summary, $\mathbf{h}^{(\text{dec})}_0, \mathbf{C}^{(\text{dec})}_0$; and a decoder $f^{(\text{dec})}$, which uses $\mathbf{h}^{(\text{dec})}_0, \mathbf{C}^{(\text{dec})}_0$ as an informed initial state to predict future observations iteratively. More precisely:
\begin{align*}
	(\mathbf{h}^{(\text{dec})}_0, \mathbf{C}^{(\text{dec})}_0) &= f^{(\text{enc})}(\mathbf{X}),\\
	\hat{\mathbf{y}}_t &= \text{Softmax}\left( f^{(\text{dec})}(\mathbf{h}^{(\text{dec})}_{t-1}, \mathbf{C}^{(\text{dec})}_{t-1}, \mathbf{x}_{t})\right)
\end{align*}
We summarise this sequence-to-sequence model in Figure \ref{fig:seq2seq}.

\subsubsection{MOrdReD: an Ordinal Regression Sequence-to-Sequence framework}\label{subsub:ordinal_seq2seq}

In our ordinal (auto-)regression setting, a sequence-to-sequence's encoder firstly produces a summary $\mathbf{h}^{(\text{dec})}_0, \mathbf{C}^{(\text{dec})}_0$ of the last $P$ observed samples of a one-hot encoded, quantised time series, $\mathbf{X}^{(\text{enc})} = (\mathbf{x}_1, \dots, \mathbf{x}_{P})$. Then, the informed summary $\mathbf{h}^{(\text{dec})}_0, \mathbf{C}^{(\text{dec})}_0$ is fed into the model's decoder, which observes the last available sample $\mathbf{x}_{P}$ and finally outputs a class probability density $\hat{\mathbf{y}}_{P} \in \mathbb{R}^M$ over the bins $C$, with $\mathbf{x}_{P+1} \sim \hat{\mathbf{y}}_P$. 

We note that $\hat{\mathbf{y}}_P$ has two possible interpretations: on the one hand, it is the categorical distribution that governs the behaviour of $\mathbf{x}_{P+1}$; on the other hand, $\hat{\mathbf{y}}_P$ itself can be interpreted as a corrupted or otherwise noisy representation of $\mathbf{x}_{P+1}$. In our model, we enable direct autoregression by feeding back $\hat{\mathbf{y}}_P$ into the decoder to forecast $\mathbf{x}_{P+2} \sim \hat{\mathbf{y}}_{P+1}$. We have observed empirically that such autoregression allows our model to perform better than by one-hot encoding either a summary statistic or a sample drawn from $\hat{\mathbf{y}}_P$. We argue that this is the case because the model has full information about its own previous decisions, including partial bin allocations.

This loopback process can be repeated an arbitrary number of times to forecast over long horizons, as we show in Section \ref{sec:experiments}. During training time, we enable teacher forcing by allowing the decoder to see the true, uncorrupted observation $\mathbf{X}_{P+k}, k>1$. In other words, we jointly train the encoder and the decoder by allowing them to see $\mathbf{X}^{(i)}, \mathbf{X}^{(i+1)}$ respectively. At testing time however, the decoder only has access to at most 1 observed sample, which it uses as a seed to perform iterative one-step ahead prediction.


\subsection{QUANTIFYING UNCERTAINTY WITH MORDRED}  \label{subsub:mc_dropout}
As described in Section \ref{subsec:nnunc}, Monte Carlo dropout can be used to approximate the predictive posterior distribution of a deep neural network. The predictive distribution of the model presented in Section \ref{subsub:ordinal_seq2seq} is given by:
\begin{align*}
	p(\mathbf{y}^* | \mathbf{X}^*) = \int p(\mathbf{y}^* | \mathbf{X}^*, \bm{\theta}) p_{\text{drop}}(\bm{\theta}) \mathrm{d}\bm{\theta},
\end{align*}
where $p_{\text{drop}}(\bm{\theta})$ denotes the dropout (Bernoulli) distribution acting as described in Section \ref{subsec:nnunc}. It is then possible to approximate the integral via Monte Carlo:
\begin{align*}
	p(\mathbf{y}^* | \mathbf{X}^*) \approx \dfrac{1}{N_s} \sum_{n=1}^{N_s} \text{Softmax} \left( f^{\hat{\bm{\theta}}^{(n)}}(\mathbf{X}^*) \right),
\end{align*}
where $\hat{\bm{\theta}}^{(n)} \sim p_{\text{drop}}(\bm{\theta})$.
Thanks to the theoretical results that connect Deep Gaussian Processes with neural networks, the use of the above mean for the predictive distribution is grounded and justified for feed-forward neural networks of arbitrary depth. These results are further extended in \citet{gal2016theoretically}, where the authors make a theoretically-sound argument for the application of dropout to the case of recurrent neural networks. In essence, the masks need to be fixed for all the recurrent passes. This is the approach we follow when performing the experiments illustrated in the remainder of this article.

We note that it is straightforward to reinterpret our ordinal predictive distribution as one over the original, pre-quantised time series. Consider a categorical output $\mathbf{\hat{y}}_t$ over the partition $C$ of intervals $C_i$, each of length $|C_i|$. Then we can straightforwardly construct a piece-wise uniform probability density function over the range of the time series $I$:
\begin{align*}
	p(\mathbf{x}_{t+1}\mid\mathbf{x}_{t},...,\mathbf{x}_{t-P+1}) &= p_{\text{MOR}}(\mathbf{x}_{t+1}\mid\mathbf{\hat{y}}_t)\\
	\nonumber
	&= \prod_{i=1}^M\frac{\mathbf{\hat{y}}_t(i)}{|C_i|}^{\mathbb{I}[\mathbf{x}_{t+1}\in C_i]}\\
	\log{p_{\text{MOR}}(\mathbf{x}_{t+1}\mid\mathbf{\hat{y}}_t)} &= \sum_{i=1}^M\pi_{t+1}^i\log{\frac{\mathbf{\hat{y}}_t(i)}{|C_i|}}\\
	\nonumber
	\pi_{t+1}^i &= \mathbb{I}[\mathbf{x}_{t+1}\in C_i]
\end{align*}
Quantities such as the negative sequence log-likelihood for a sequence of $P_h$ samples drawn from the pre-quantised time series can now be calculated as:
\begin{align*}
	\text{NLL}_{\text{MOR}} 
	&= -\log{\prod_{k=1}^{P_h}  \prod_{i=1}^M\frac{\mathbf{\hat{y}}_{N+k-1}(i)}{|C_i|}^{\pi_{N+k}^i}}\\
	\nonumber
	&=-\sum_{k=1}^{P_h}\sum_{i=1}^M\pi_{N+k}^i\log\frac{\mathbf{\hat{y}}_{N+k-1}(i)}{|C_i|}
\end{align*}

\subsection{REMARKS}
Two crucial aspects to highlight about our framework are the following. Firstly, we achieve end-to-end time series forecasting by allowing our model to learn directly from data. Importantly, we do not make of use of handcrafted features, nor require expert knowledge to specify the core parts of the model (as often required in GP kernel design for instance).

Secondly, our model learns a non-parametric predictive posterior at each time step. This allows for rich, multi-modal behaviour to be encoded and fed back in an autoregressive fashion. We have observed empirically that providing our model with full knowledge of its own previous decisions enables it to achieve more reliable, longer-term forecasting than by one-hot encoding some statistic (such as the empirical mean or the mode) at every time step.




\subsection{BASELINE MODELS}\label{sub:baselines}

In this Section, we review the baseline models we use to compare our framework with. These largely belong to three different families:

\begin{itemize}
    \item \textbf{RNN regression.} This baseline is a simple counterpart of our model that motivates the transition from direct to ordinal regression with RNNs. The same sequence-to-sequence architecture described in Section \ref{subsub:seq2seq}, but learnt by minimising the mean squared error of the model output with respect to the raw time series.
    \item \textbf{AR(p) modelling formulated as a state-space model.} Autoregressive modelling of order $p$ is a well-established statistical method, and widely used as a benchmark for time series forecasting tasks. We simply recapitulate that AR(p) models with time-varying coefficients can be recast as linear Gaussian state-space models. Such systems admit exact parameter inference and prediction in an efficient and principled manner by means of the Kalman filter. We refer the reader to \cite{durbin2012time} for a more detailed presentation of this model.
    \item \textbf{Gaussian Process autoregression.} GPs, and in particular autoregressive GPs, have dominated the time series literature over the last decade. One potential limitation of this approach is that iterative one-step forecasting indeed requires GPs able to handle noisy inputs, which is still an active topic of research \citep{girard2005gaussian}. We thus propose two autoregressive GP baselines: one in which the one-step-ahead predictive posterior distributions are approximated via Monte-Carlo integration (described in Section \ref{subsub:gp_mc}), and an alternative approach in which each predictive posterior is approximated with a Bayesian Gaussian Mixture Model (described in Section \ref{subsub:gp_gmm}). The latter approach requires more computational resources, but also enables the model to represent rich, multi-modal behaviour and adapt it over time.
\end{itemize}

\subsubsection{Autoregressive Gaussian Processes} \label{subsub:gp_mc}

Gaussian processes describe distributions over functions. They are defined as a collection of random variables, any finite number of which have a joint Gaussian distribution. A GP $f$ is fully specified by a mean function $m(\mathbf{x}) = \mathbb{E}[f(\mathbf{x})]$ and a covariance function $k(\mathbf{x}, \mathbf{x}) = \mathbb{E}[(f(\mathbf{x}) - m(\mathbf{x}))(f(\mathbf{x}') - m(\mathbf{x}'))]$. The covariance function (or kernel) describes correlations between data points and can be used to encapsulate prior belief about the problem. We refer the reader to \citet{Rasmussen2006} for a more detailed discussion about Gaussian Processes.

Consider a time series of length $N$. Long-term forecasting can be done in an autoregressive manner by learning a Gaussian Process $f$ that maps a sequence of length $P$, $\mathbf{X} = (\mathbf{x}_{N-P-1}, \dots, \mathbf{x}_N)$, to a normal distribution over its next sample $\mathcal{N}(\mathbf{x}_{N+1}\mid\bm{\mu}_{N+1}, \bm{\sigma}^2_{N+1})$. However, sequentially inferring $\bm{\mu}_{t}, \bm{\sigma}^2_{t}$ for any $t>N+1$ requires adequate forward-propagation of the uncertainty computed in all previous forecasts. This effectively turns the task into one of GP regression with noisy inputs.



Uncertain, or noisy input, GPs are analytically intractable and therefore developing approximations remains an active area of research. In our work, we use a Monte Carlo-based approach \citep{girard2005gaussian}. $S_{\text{GP}}$ sample trajectories are drawn up to time $N+P_h$ for a predictive horizon $P_h$, i.e. $\hat{\mathbf{X}}^{[s]} = (\hat{\mathbf{x}}^{[s]}_{N+1}, \dots, \hat{\mathbf{x}}^{[s]}_{N+P_h})$, and used to estimate the moments ${\tilde{\bm{\mu}}_{N+k}}$, ${\tilde{\bm{\sigma}}^2_{N+k}}$ of the corrected normal distribution $p(\mathbf{x}_{N+k} \mid \tilde{\bm{\mu}}_{N+k}, \tilde{\bm{\sigma}}^2_{N+k})$.

Each sample trajectory is built iteratively $\forall k, 2\leq k\leq P_h$ in the following fashion: 
\begin{enumerate}
	\item at time $N+k-1$, compute $\mathcal{N}(\mathbf{x}_{N+k-1}\mid$ $\bm{\mu}_{N+k-1},$ $\bm{\sigma}^2_{N+k-1})$ from the sequence $\hat{\mathbf{X}}^{(N+k-2)}$,
	\item draw a sample $\hat{\mathbf{x}}^{[s]}_{N+k-1}$ from $\mathcal{N}(\mathbf{x}_{N+k-1} \mid$  $\bm{\mu}_{N+k-1},$ $\bm{\sigma}^2_{N+k-1})$, and build the next input seed sequence 
	\begin{align*}
	    \hat{\mathbf{X}}^{(N+k-1)} = (\mathbf{x}_{N+k-P-1}, \dots, \mathbf{x}_{N+k-1})
	\end{align*}
	\item repeat from step 1 until $k>P_h$.
\end{enumerate}
After repeating the procedure above $S_{\text{GP}}$ times, the corrected mean and variance $\forall k, 2\leq k \leq P_h$ are thence given by:
\begin{align*}
	\tilde{\bm{\mu}}_{N+k} &= \mathbb{E}[\mathbf{x}_{N+k}]
	\approx \frac{1}{S_{\text{GP}}}\sum_{s=1}^{S_{\text{GP}}}{\hat{\mathbf{x}}^{[s]}_{N+k}}\\
	\tilde{\bm{\sigma}}^2_{N+k} &= \mathbb{E}\left[ (\mathbf{x}_{N+k})^2 \right] - \mathbb{E}\left[\mathbf{x}_{N+k}\right]^2 \\
	\nonumber
	&\approx \frac{1}{S_{\text{GP}}}\sum_{s=1}^{S_{\text{GP}}}{\left(\hat{\mathbf{x}}^{[s]}_{N+k} - \tilde{\bm{\mu}}_{N+k} \right)^2}
\end{align*}

\subsubsection{Bayesian Gaussian Mixture Models}  \label{subsub:gp_gmm}
The baseline described above assumes each one-step-ahead predictive distribution is uni-modal Gaussian. We propose an alternative baseline in which we characterise the sample trajectories just described as a collection of Gaussian Mixture Models (GMM), which allows for one-step-ahead predictive distributions with a variable number of modes at each time step. In other words, the prediction $\mathbf{x}_{t+1}\sim\text{GMM}(\mathbf{x}_{t+1}\mid\bm{\pi}_{t+1},\bm{\mu}_{t+1},\bm{\sigma}^2_{t+1})$ with
\begin{align*}
    \begin{multlined}
    \text{GMM}(\mathbf{x}_{t+1}\mid\bm{\pi}_{t+1},\bm{\mu}_{t+1},\bm{\sigma}^2_{t+1}) =\\ \sum_{k=1}^K \bm{\pi}_{k, t+1} \mathcal{N}(\mathbf{x}_{t+1}\mid\bm{\mu}_{k, t+1},\bm{\sigma}^2_{k, t+1})
\end{multlined}
\end{align*}
where each $\bm{\pi}_{k, t+1}$ is the mixture weight of the individual Gaussians and $K$ is a hyperparameter for the number of mixtures in the model. Indeed this enables the description of richer behaviour than the one given by uni-modal Gaussians, but comes at the expense of the potentially computationally expensive process of learning up to $P_h$ distinct GMMs.

There is extensive literature available on the inference of the parameters $(\bm{\pi}_{t+1},\bm{\mu}_{t+1},\bm{\sigma}^2_{t+1})$, and indeed on model selection for the appropriate number of mixtures $K$. State-of-the-art methods however aim to perform Bayesian inference over the learnable parameters, i.e. inferring a posterior distribution over each learnable parameter that best explains the observed evidence or data. This enables a principled prediction framework for other quantities of interest and further guards against overfitting and other well-documented pathologies that occur in point-estimate inference mechanisms, such as Maximum a Posteriori estimation \citep{Bishop2006}. Furthermore, such Bayesian methods also allow for natural shrinkage of the number of mixtures $K$, i.e. the weights of less informative mixtures naturally shrink to $0$ and thus one need only provide an upper bound on $K$. We note that, although full Bayesian inference is intractable, efficient approximate Bayesian inference can be performed via Variational Bayes techniques, which is the approach we take in this article. We refer the reader to \cite{penny2000bayesian}, \cite{bernardo2003variational}, \cite{ghahramani2001introduction} and \cite{Murphy2012} for a more detailed discussion of Variational Bayes GMMs.





	\section{EXPERIMENTS}
\label{sec:experiments}

In this Section, we discuss our model's performance in two different long-term different tasks following the presentation of our baselines in Section \ref{sec:data}.

\subsection{DATA}
\label{sec:data}
Our large-scale experiment was performed on 45 datasets compiled from different sources \citep{Fulcher20130048,rezek1998stochastic,Bramblemet}. We used the following criteria to obtain this compilation:

\begin{itemize}
    \item \textbf{Length}. The compilation includes time series of different lengths, mostly ranging between 1,000 and 20,000 samples. In this work, we aim to compare our model in a setting in which poor performance should not be attributed to lack of data. We therefore only use time series with at least 10,000 observations, and further truncate those that contain over 30,000 so as to retain computational tractability during model fitting.
    \item \textbf{Structure}. All time series were drawn from either the provided real-world references, or are a synthetic dynamical system.
    \item \textbf{Domain variety}. Time series were queried and grouped by application domain, and at most two from each resulting group were subsequently drawn at random.
\end{itemize}

The selected datasets correspond with a wide range of dynamical systems, chaotic maps, and data obtained from varied application domains, including meteorology, astronomy and anatomy. A summary list and short descriptions of the selected datasets is given in Appendix A, and full descriptions can be found in the supplementary material of \citep{Fulcher20130048,rezek1998stochastic,Bramblemet}. All the time series in the library were linearly and seasonally de-trended as detailed in the corresponding sources. Furthermore, all time series were standardised to zero mean, unit variance.

\subsection{TASK 1: LONG-TERM FORECASTING}\label{sub:uncer}
In this experiment, we perform a large-scale comparison of our MOrdReD framework against the 4 baselines described in Section \ref{sub:baselines}. Our long-term forecasting task consists in extrapolating $P_h$ emissions $(\hat{\mathbf{x}}_{N+1}, \dots, \hat{\mathbf{x}}_{N+P_h})$ forward from the last $P$ observations $(\mathbf{x}_{N-P}, ..., \mathbf{x}_N)$ in the time series of length $N$, with $P<<P_h$. In our setting, we chose a lookback $P=100$, and a predictive horizon $P_h=1000$. Such iterative prediction is obtained for each of the 45 datasets described in Section \ref{sec:data} without looking at real testing data or re-estimating model parameters during the process. 

We note that the aim of this forecasting task is to showcase the performance of our method as an ``out-of-the-box" methodology, and thus we perform our comparison prior to any application domain crafting takes place (such as tailoring a task-optimal kernel, or appending handcrafted features and exogenous or correlated time series). We now give specific details about data preparation, model selection, model implementation and evaluation metrics.

\subsubsection{Data pre-processing}  \label{subsub:data_prep}
We used 45 different datasets drawn from both synthetic and real-world sources. Synthetic data includes an ample number of dynamical systems and chaotic maps, and real-world data was drawn from a number of sources and application domains, which encompass Meteorology, Astronomy, Physiology, Acoustics, and others. A short list is provided in Appendix A, and a more detailed description is given in \cite{Fulcher20130048}. All datasets were linearly and seasonally de-trended as detailed in Section 2.6 of the complementary material of \cite{Fulcher20130048}. Furthermore, all time series were standardised to zero mean, unit variance.  In the case of MOrdReD, all time series were quantised into a maximum of $M=300$ bins. For the rest of the models, which perform direct regression, regularising white noise with $\sigma=10^{-3}$ was added to the standardised time series. Datasets were further split into training, validation and testing time series. The training sets correspond to the first 70\% of the observations in each time series, validation sets correspond to the subsequent 15\%, and test sets to the last 15\%. 

\subsubsection{Model learning, selection and implementation}  \label{subsub:model_selec}
We now provide details for the optimisation of each model's parameters and hyperparameters. The optimal hyperparameter configurations found for our neural network and AR(p) models are given in Appendix C.

\begin{itemize}
    \item \textbf{MOrdReD.} All our MOrdReD models were built in Python 2.7 and using Keras 2.1.2 with the Tensorflow backend \citep{chollet2015keras}. Parameter optimisation was performed by minimising the categorical cross-entropy with Nesterov momentum Adam using $\alpha=0.002, \beta_1=0.9, \beta_2=0.999$ and a schedule decay of $0.004$, and parameters were initialised using the Glorot Uniform method. At training time, all models incorporate Early Stopping to guard against overfitting, while allowing for a maximum of 50 epochs using mini-batches of 256 training examples. Both the bidirectional encoder and the decoder share the same number of hidden units $n_u\in\{64, 128, 256, 320\}$, and the output fully-connected layers have $M$ softmax output units corresponding to each time series' ordinal classes. All models further incorporate dropout and $L2$ regularisation with hyperparameters $p_{\text{dout}} \in \{0.25, 0.35, 0.5\}$ and $\lambda_{L2}\in\{1e-6, 1e-7, 1e-8\}$, which were jointly optimised with the number of hidden units $n_u$ by hypergrid search on the validation set. MC dropout predictive posteriors were estimated using $N_s=100$ samples.
    \item \textbf{Direct regression sequence-to-sequence neural network.} These were learnt in a similar fashion as the above, but by minimising the mean squared error loss function of the single output unit (with no activation function) with respect to the ground truth. This baseline serves to motivate the transition into ordinal regression.
    \item \textbf{Autoregressive GP.} All our GP models were implemented using GPy 1.9.2 \citep{gpy2014} and using a Mat\'ern 5/2 kernel with additive white noise and Automatic Relevance Determination enabled. Samples drawn from the Mat\'ern 5/2 covariance function are twice-differentiable, which enables the modelling of a wide range of physical systems such as those in our datasets. Time-indexed Gaussian predictive distributions were then estimated following the Monte Carlo-based procedure described in Section \ref{subsub:gp_mc}, with $N_s=100$ samples trajectories.
    \item \textbf{ARGP with predictive Gaussian Mixture Model.} This baseline follows the implementation described above, but the time-indexed predictive distributions are estimated via a Gaussian Mixture Model as described in Section \ref{subsub:gp_gmm}. This allows for rich, multi-modal behaviour to be described succinctly at each time step. We performed approximate Bayesian inference to learn these models, with up to $K=5$ mixtures at each time step, using the open source Variational Bayes GMM implementation available in the Scikit-learn toolbox \citep{scikit-learn}, v. 0.19.1.
    \item \textbf{AR(p).} The Python library StatsModels 0.9 \citep{seabold2010statsmodels} offers an open source implementation of the state-space formulation of AR(p) modelling. We used this and optimised the lookback hyperparameter $p\in\{16, 32, 64\}$.
\end{itemize}

Our experiments are further accompanied by a Python library that provides an implementation of our methodology, as well as an interface for the ample number of libraries that implement our baselines. This code and the scripts used to perform these experiments can be found at \url{https://www.github.com/bperezorozco/ordinal_tsf}.

\subsubsection{Evaluation metrics}  \label{subsub:metrics}
Our evaluation is performed across three axes: predictive accuracy, uncertainty quantification and forecast reliability. This exhaustive comparison enables us to speak not only about the performance of the forecasts predicted by our model, but also about the credibility and reliability of the uncertainty bounds they infer. In terms of forecast accuracy, we measure the Symmetric Mean Absolute Percentage Error (SMAPE) and the Root Mean Squared Error (RMSE) of both the mean and the median of the predictive distributions of each method:
\begin{align*}
    \text{SMAPE} &= \frac{2}{P_h}\sum_{t'=N}^{N+P_h}{\frac{|\mathbf{x}_{t'} - \mathbf{\hat{x}}_{t'}|}{|\mathbf{x}_{t'}| + |\mathbf{\hat{x}}_{t'}|}} 
    \\
    \text{RMSE} &= \sqrt{\frac{\sum_{t'=N}^{N+P_h}{\lVert\mathbf{x}_{t'} - \mathbf{\hat{x}}_{t'}\rVert^2}}{P_h}}
\end{align*}

Crucially, one goal in this experiment is not just to evaluate a model's predictive accuracy, but also to measure its honesty expressed through uncertainty estimations. We prefer models that are able to either produce an accurate confident forecast, or otherwise explicitly confess their ignorance - for instance, through a mean-reversal process. 

We quantify this via the negative sequence log-likelihood (NLL), which we rewrite using the chain rule for joint distributions:
\begin{align*}
    -\log{p\left(\mathbf{X}^{(N+P_h)} \right)} = &-\log{p(\mathbf{x}_{N+1})} \\ &-\sum_{k=2}^{P_h}\log{{p\left(\mathbf{x}_{N+k}\Bigg\vert \bigcap_{k'=1}^{k-1}\mathbf{x}_{N+k'}\right)}}
\end{align*}

For MOrdReD, each term in the summation acquires the piece-wise uniform shape described in Section \ref{subsub:mc_dropout}; for ARGPs with a GMM predictive distribution, it acquires the shape described in Section \ref{subsub:gp_gmm}; for all the other baselines, it is given by a standard Gaussian density. 

In addition to the NLL, we also provide the integrated or cumulative NLL:

\begin{align*}
    \text{CNLL} = \sum_{i=1}^{P_h}{-\log{p\left(\mathbf{X}^{(N+i)} \right)}}
\end{align*}

This metric summarises information about how the NLL changes with time, e.g. greater penalties are incurred by models that make inaccurate short-term predictions, whereas erroneous long-term forecasts have a lesser contribution.

Lastly, we benchmark our methodology in terms of the calibration of its uncertainty bound predictions. That is, we measure its ability to produce output densities that can be interpreted as real-world probabilities. Graphics tools such as quantile-quantile (QQ) plots and reliability diagrams are widespread in the time series literature to measure uncertainty calibration. However, due to the large-scale nature of this task, we instead propose a related summary metric of these which we now introduce. 

Consider the time series $\mathbf{\hat{x}}_{\alpha}$ whose $t$-th entry is given by the $\alpha$-quantile of the predictive distribution at time $t$, i.e.:
\begin{align*}
    \mathbf{\hat{x}}_{\alpha,t} = q_{t} \qquad \text{s.t.} \bigintss_{-\infty}^{q_t}{p\left(\mathbf{x}_{t}\Bigg\vert \bigcap_{t'=1}^{t-1}\mathbf{x}_{t'}\right)d\mathbf{x}_t} = \alpha
\end{align*}

For a given calibrated model and $\alpha, 0 < \alpha < 1$, we expect that exactly $\alpha$\% of the observations in the ground truth $\mathbf{x}$ will lie below $\mathbf{\hat{x}}_{\alpha}$. Let $r_\alpha$ be this proportion. Then for better calibrated models, $r_\alpha \rightarrow \alpha$ with:

\begin{align*}
    r_\alpha = \frac{1}{P_h}\sum_{t=N+1}^{N+P_h}{\mathbb{I}[\mathbf{x}_{t} < \mathbf{\hat{x}}_{\alpha, t}]} 
\end{align*}

We thus define the calibration metric QQDist as the (Euclidean) distance between $r_\alpha$ and $\alpha$:
\begin{align*}
    \text{QQDist} = \int_{0}^1{(r_\alpha - \alpha)^2d\alpha}
\end{align*}

We provide two instances of this metric. One in which the $r_\alpha$ for each method are computed for the full predictive horizon $P_h=1000$, and one up to a truncated horizon $P'_h=250$. We provide both instances so as to distinguish calibration performance in both the medium and long term.

Performance is measured over the following 6 statistics: SMAPE and MASE for the predictive median's accuracy; NLL and CNLL to evaluate the predictive distributions' accuracy; and QQDist and QQDist-250 for uncertainty calibration.

\subsubsection{Results}\label{subsub:results}
We summarise our results in Tables \ref{tab:results_1}, \ref{tab:results_3} and \ref{tab:results_4}. We note that our model achieves state-of-the-art performance in an wholly end-to-end fashion, with no human intervention to specify the core architecture of the model. Unlike GP Regression, where knowledge of kernels is required and often inaccessible to unfamiliar users, our framework is end-to-end and  requires almost no user intervention. Additionally, other advantages of neural networks such as scalability are now readily accessible. Our framework can make the most out of big datasets and produce reliable forecasts in the long term that outperform other state-of-the-art techniques.

From Table \ref{tab:results_1} we note that our model is the one that performs best in the largest number of datasets across all the proposed metrics. Interestingly, we see that the worst performing baseline is the one given by direct regression neural networks. This gives further empirical motivation to transition from direct into ordinal regression for the family of neural network models. We also note that classical statistical models perform very well in the calibration metrics, whereas GPs are the worst performing in this axis. We argue that this could be the case due to the exact sequential inference method of the predictive variance proposed in the state-space formulation of AR(p) models. Finally, GPs are the closest competitors in terms of forecasting accuracy.

We note that, for some time series, our model is the only one capable of learning any structure for long-term forecasting. One such example is the electrocardiogram time series, depicted in Figure \ref{fig:heart}. This complex-shaped signal is consistently predicted in a timely and accurate fashion by MOrdReD in the long-term, whereas no baseline is capable of reproducing the characteristic shape of the QRS complex beyond its first occurence.

We now focus on Table \ref{tab:results_3}, which provides the mean performance rank for each baseline and metric. This provides further information about how our method performs when it is not the best-performing one, and we observe that MOrdReD still achieves the best mean rank for all metrics. In order to produce this table, two considerations were taken into account: on one hand, both GP baselines could propose very similar predictive distributions for some datasets (e.g. those in which multi-modality is not required and therefore both propose a uni-modal Gaussian), and this induced twice the penalty for those methods that underperformed against GPs. We solve this by merging the results of both GPs and retaining the better performing one. This avoids allocating a double penalty on models that underperform with respect to GPs if both GPs propose equivalent predictive distributions (e.g. if the optimal number of Gaussian mixtures is 1).

The information in Table \ref{tab:results_3} crucially summarises that whenever our model is not the best performing, it is still likely to be the second best-performing, etc. An example of this scenario is given in Figure \ref{fig:tide}, which provides our forecasts for the tide height dataset. The GP baseline is the best performing, as it achieves better timing accuracy in the longer term in comparison with MOrdReD. However, the remaining baselines are not able to capture the basic structure of the signal at all, and therefore our model comes out as a clear runner-up.

\begin{figure*}
		\centering
		\includegraphics[width=5in,height=8in]{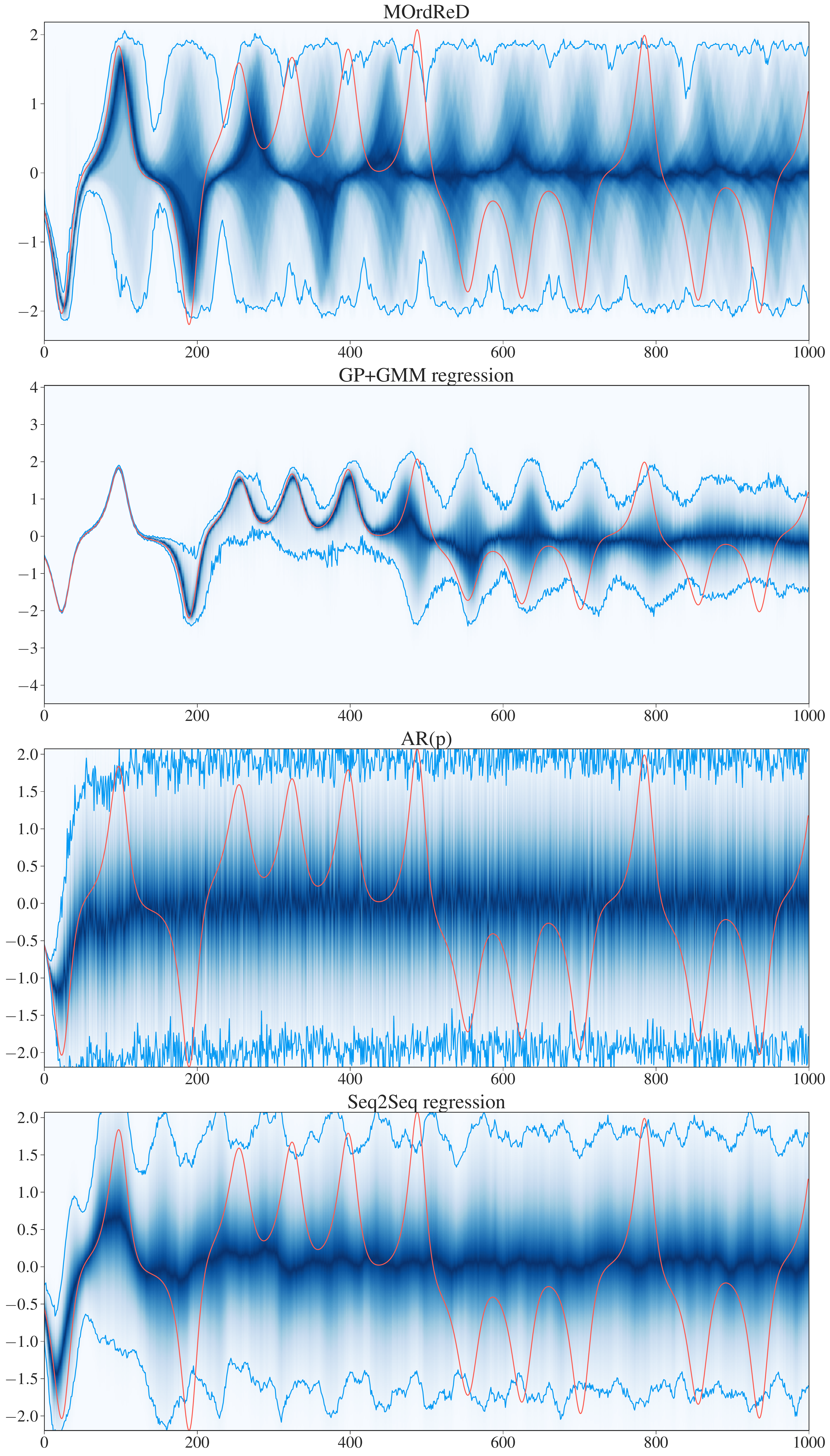}
	\caption{Example of out-of-sample forecast with predictive horizon $P_h=1000$ for the Lorenz chaotic map time series dataset. Both MOrdReD and GP+GMM regression are capable of modelling adaptable multi-modal behaviour through time, but our model phases out with respect to the ground truth before the GP does. This enables the baseline the achieve a clear performance advantage with respect to our model in this case. Ground truth (orange) overlayed on top of the predictive distributions given by each model. 95\% confidence bounds are highlighted in light blue.}
	\label{fig:lorenz}
\end{figure*}

Nevertheless, we also observed that MOrdReD is not always so close to the best-performing model. For instance, we noted that GPs are often the best-performing in chaotic maps and dynamical systems. An example is given in Figure \ref{fig:lorenz}, where we provide the forecasts for a Lorenz map. Whereas both MOrdReD and GPs are able to model multi-modal behaviour in the long-term, MOrdReD syncs out with respect to the ground truth considerably faster than GPs. Indeed, GPs only make use of their multi-modal predictive capability relatively late in comparison with MOrdReD.

In a similar fashion to Table \ref{tab:results_3}, in Table \ref{tab:results_4} we introduce the reciprocal metric: mean \textit{worst-performance rank}. This metric speaks about the frequency in which a method is the worst performing for a given task. Rank 3 for each task in this case is given to the best-performing model, and thus larger overall metric values indicate better performance. These results indicate that there are relatively few cases in which all other models outperform it for any given metric in the assessment.

\begin{table*} \centering
\begin{tabular}{lrrrrrrrr}
\toprule
{} &     NLL &  CumuNLL &  Mean RMSE &  Med RMSE &  Mean SMAPE &  Med SMAPE &  QQ Dist &  QQ Dist 250 \\
\midrule
\textbf{MOrdReD           } & \textbf{22} &         \textbf{22} &    \textbf{17} &      \textbf{11} &         \textbf{15} &        \textbf{22} &  \textbf{18} &      \textbf{18} \\
\textbf{GPR     } &  3 &          3 &     0 &       7 &          0 &         4 &   0 &       5 \\
\textbf{GPR+GMM } &  9 &         11 &    10 &      10 &         \textbf{15} &        10 &   6 &       2 \\
\textbf{AR(p)             } &  9 &          6 &    10 &       8 &          7 &         3 &  \textbf{18} &      15 \\
\textbf{Seq2Seq Reg.} &  2 &          3 &     8 &       9 &          8 &         6 &   3 &       5 \\
\bottomrule
\end{tabular}
\caption{Number of datasets in which the each achieves the best performance amongst all competitors. \textbf{Bold} indicates best performance. We provide full details of these results in Appendix B.}
\label{tab:results_1}
\end{table*}
~
\begin{table*} \centering
\begin{tabular}{lrrrrrrrr}
\toprule
{} &  NLL &  CumuNLL &  Mean RMSE &  Med RMSE &  Mean SMAPE &  Med SMAPE &  QQ Dist &  QQ Dist 250 \\
\midrule
\textbf{MOrdReD           } & \textbf{0.80} &            \textbf{0.78} &       \textbf{1.20} &         \textbf{1.33} &            \textbf{1.18} &           \textbf{0.84} &     \textbf{1.13} &         \textbf{1.16} \\
\textbf{Best GP           } & 1.44 &            1.20 &       1.36 &         1.36 &            1.38 &           1.40 &     1.58 &         1.82 \\
\textbf{AR(p)             } & 1.44 &            1.76 &       1.58 &         1.78 &            1.73 &           1.87 &     1.33 &         1.20 \\
\textbf{Seq2Seq Reg} & 2.31 &            2.27 &       1.87 &         1.53 &            1.71 &           1.89 &     1.96 &         1.82 \\
\bottomrule
\end{tabular}
\caption{Average performance rank achieved by each model for each metric, where rank 0 is given to the \textit{best} performing model and 3 to the worst performing. Best performance in given in \textbf{bold}. We note that MOrdReD consistently performs best in this metric, and in particular that the expected scenario is that it will be either the best or second-best performing model for a given task.}
\label{tab:results_3}
\end{table*}
~
\begin{table*}[!ht] \centering
\begin{tabular}{lrrrrrrrr}
\toprule
{} &  NLL &  CumuNLL &  Mean RMSE &  Med RMSE &  Mean SMAPE &  Med SMAPE &  QQ Dist &  QQ Dist 250 \\
\midrule
\textbf{MOrdReD           } & \textbf{2.31} &            \textbf{2.27} &       \textbf{1.87} &         1.53 &            1.71 &           \textbf{1.89} &     \textbf{1.93} &         1.80 \\
\textbf{Best GP           } & 1.44 &            1.76 &       1.58 &        \textbf{ 1.78} &            \textbf{1.73} &           1.87 &     1.36 &         1.22 \\
\textbf{AR(p)             } & 1.44 &            1.20 &       1.36 &         1.36 &            1.38 &           1.40 &     1.58 &         \textbf{1.82} \\
\textbf{Seq2Seq Reg} & 0.80 &            0.78 &       1.20 &         1.33 &            1.18 &           0.84 &     1.13 &         1.16 \\
\bottomrule
\end{tabular}
\caption{Average worst performance rank achieved by each model for each metric, where rank 0 is given to the \textit{worst} performing model and 3 to the best performing. Best performance in given in \textbf{bold}. We note that MOrdReD either performs best or is the runner-up in this metric, and in particular we note that the expected scenario is that there will be very few cases in which \textit{all} baselines will outperform it.}
\label{tab:results_4}
\end{table*}

\subsection{TASK 2: QUANTIFYING EVENT TIMING UNCERTAINTY}\label{sub:timing}
As motivated earlier, a crucial component of long-term prediction is \textit{timely} forecasting. Decision-makers often prioritise knowing the time of occurrence of an event over the magnitude of the event itself. For instance, consider the tide height dataset of Section \ref{sec:data}. Experts are often interested in knowing \textit{when} the tide levels will reach their maximum, and not just the actual height they will reach.\footnote{Despite this being a periodic phenomenon, sensor quantisation leads to representation errors that make the time series quasi-periodic in practice.}

In a similar vein, cardiologists are interested, for instance, in forecasting the timing of events such as the QRS complex. Predicting these timings further enables them to compute other metrics of interest, such as the RR interval -- which has been linked, for example, to Parkinson's disease. Accurately quantifying these metrics is therefore of utmost importance, and fully relies on models that can speak honestly about their forecasts and their uncertainties.

 The task we now focus on is to predict when a certain event will happen, alongside reliable uncertainty bounds. In this setting, we compare our MOrdReD framework with the same baselines described in Section \ref{sub:baselines}. However, in contrast to Task 1 in Section \ref{sub:uncer}, we focus our attention on only 8 datasets drawn from meteorology and physiology. We motivate our decision as follows: on one hand, evaluating the occurrence of local maxima is not relevant for some data (e.g. acoustic signals); on the other, we wish to focus our attention on real-world data, subject to realistic noise. 

\subsubsection{Constructing densities for critical event occurrence}\label{subsub:exp2}

In order to adequately quantify event timings, we construct a non-parametric probability distribution $p(t)$ that describes the probability of an event happening at time $t$. In our setting, we take, as example, this event to be the time series hitting a local maximum. To compute such timing forecasts, we employ Kernel Density Estimation (KDE) to construct a probability density function $p(t)$ that describes the probability that a certain event (hitting a local maximum in this example) will happen at time $t$. KDE is a smoothing technique that builds a non-parametric distribution from a sample set $X$. The estimator is given by

\begin{align*}
	p(t) = \frac{1}{n}\sum_{i=1}^{|X|}K_h(t-t_i),
\end{align*}
where $t_i\in X$, $K_h$ is a non-negative function that integrates to one, and $h$ is the bandwidth hyperparameter. The reader is referred to \citet{hastie01statisticallearning} for a more detailed discussion. In our work, we used the freely available implementation of Scikit-learn \citep{scikit-learn}. Now consider the true event timings $X^{\text{(true)}} = (t_1^{(\text{true})}, \dots, t_{L'}^{(\text{true})})$. The constructed densities are then evaluated in terms of the negative log-likelihood for effectively predicting such correct timings:
\begin{align*}
\text{NLL}_{X^{\text{true}}} = -\log{\prod_{i=1}^{L'}p\left(t_i^{(\text{true})}\right)} = -\sum_{i=1}^{L'}\log{p\left(t_i^{(\text{true})}\right)}
\end{align*}

We now briefly describe how to construct the sets $X, X^{(\text{true})}$. In the case of $X$, sample trajectories are drawn from the predictive densities of each framework, and the timing of the desired event in each sample is recorded. An estimator $p(t)$ is subsequently built for each method. In the case of $X^{(\text{true})}$, the true event timings are recorded directly from the ground truth. Maxima detection was achieved in our experimental setting with the open source PeakUtils Python library. Recording the timing of events can be inhibited by the presence of noise, even in the ground truth data. To construct the set $X^{(\text{true})}$, we decomposed each dataset's ground truth using Empirical Mode Decomposition \citep{huang1998empirical} and then recorded the events from the (noiseless) intrinsic mode function that captured long-term quasi-periodic nature. To this end, we used an additional Python open source library, PyEMD.

\subsubsection{Results}\label{subsub:results2}

\begin{table*}
	\centering
	\begin{tabular}{lrrrr}
    \toprule
    {} &  MOrdReD &  Best-performing GP &   AR(p) &  Seq2Seq regression \\
    \midrule
    \textbf{CM\_air 1} &  \textbf{19.4725} &  21.1712 & 20.3755 &             22.1955 \\
    \textbf{CM\_air 2  } &  22.7064 &  \textbf{18.3212} & 20.2896 &             21.0814 \\
    \textbf{CM\_rhum   } &  \textbf{20.5574} &  20.6010 & 20.6663 &             21.1996 \\
    
    \textbf{CM\_slp 1        } &  14.1069 &  \textbf{13.8725} & 14.0731 &             14.0154 \\
    \textbf{CM\_slp 2   } &  \textbf{12.1217} &  13.2720 & 13.7170 &             13.8411 \\
    \textbf{AIRFLOW   } & \textbf{ 12.1408} &  12.5615 & 14.6095 &             13.3766 \\
    \textbf{ECG     } &  \textbf{68.6775} &  81.6095 & 82.6156 &             81.2865 \\
    \textbf{TIDE    } &  32.7297 &  \textbf{32.3702} & 41.3097 &             41.0631 \\
    \midrule
     \textbf{\# BEST} &  \textbf{5} &  3 & 0 &             0\\
    \bottomrule
    \end{tabular}
    \caption{Negative log-likelihood for the event detection task. Figures in \textbf{bold} indicate best performance.}
    \label{tab:timing}
\end{table*}

In Table \ref{tab:timing}, the negative log-likelihood values for drawing the true timings $X^{(\text{true})}$ from the constructed densities $p_{\text{ARGP}}(t),$ $p_{\text{MOR}}(t)$ are shown for predicting the occurrence of a local maximum in 8 different datasets. We observed that our framework does best in more dataset examples than the rest of the baselines, closely followed by the GP baseline, which dominates in the remaining datasets. 

\begin{figure*}
		\centering
		\includegraphics[width=5in,height=8in]{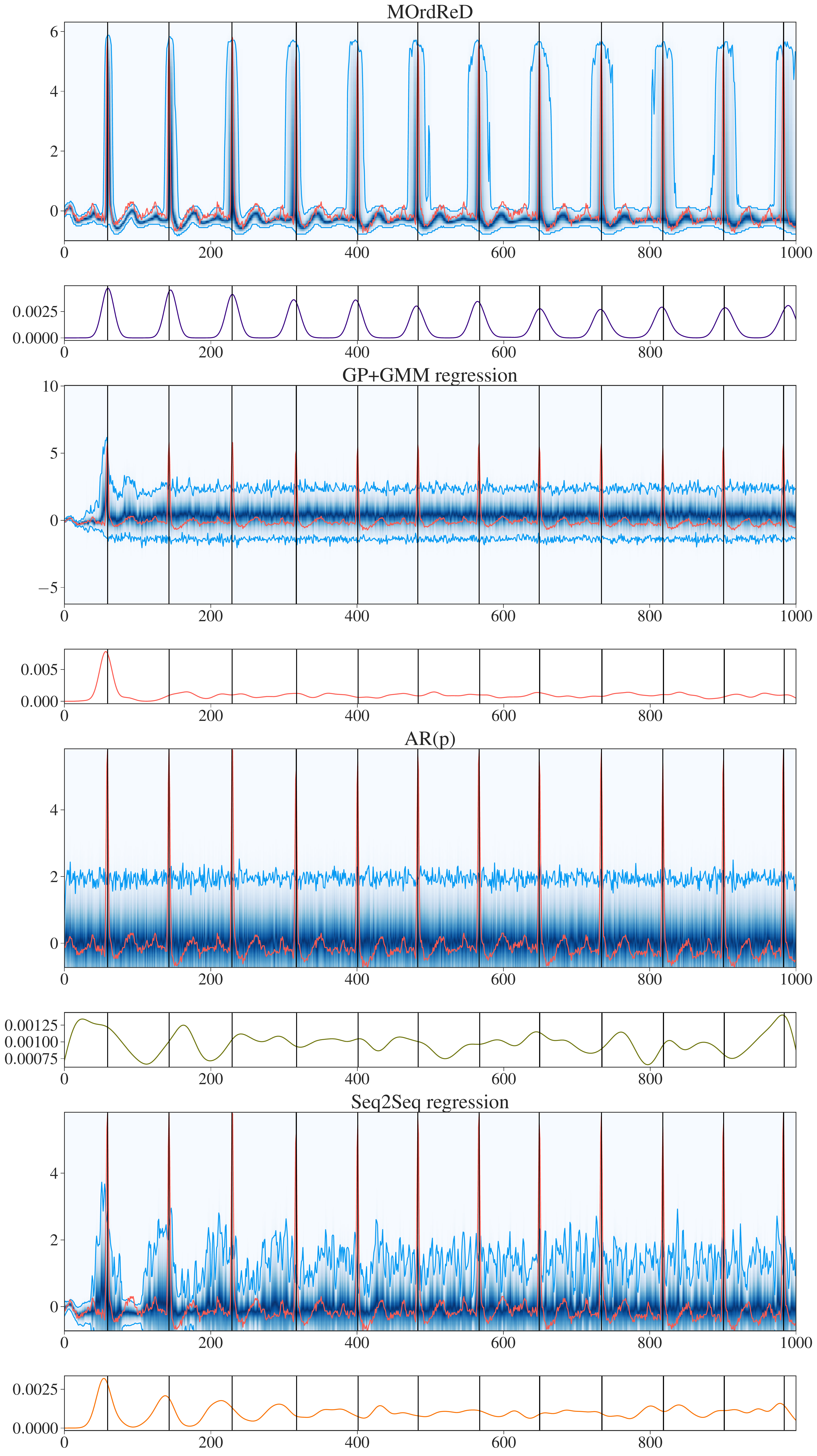}
	\caption{Example of out-of-sample forecast with predictive horizon $P_h=1000$ for an electrocardiogram time series dataset. This is an example of a complex-shaped signal that is consistently predicted in an accurate in timely fashion by our framework only. Ground truth (orange) overlayed on top of the predictive distributions given by each model. 95\% confidence bounds are highlighted in light blue. Maxima predictive densities are provided under each predictive distribution.	Vertical lines were drawn at the optima of interest.
	}
	\label{fig:heart}
\end{figure*}

\begin{figure*}
		\centering
		\includegraphics[width=5in,height=8in]{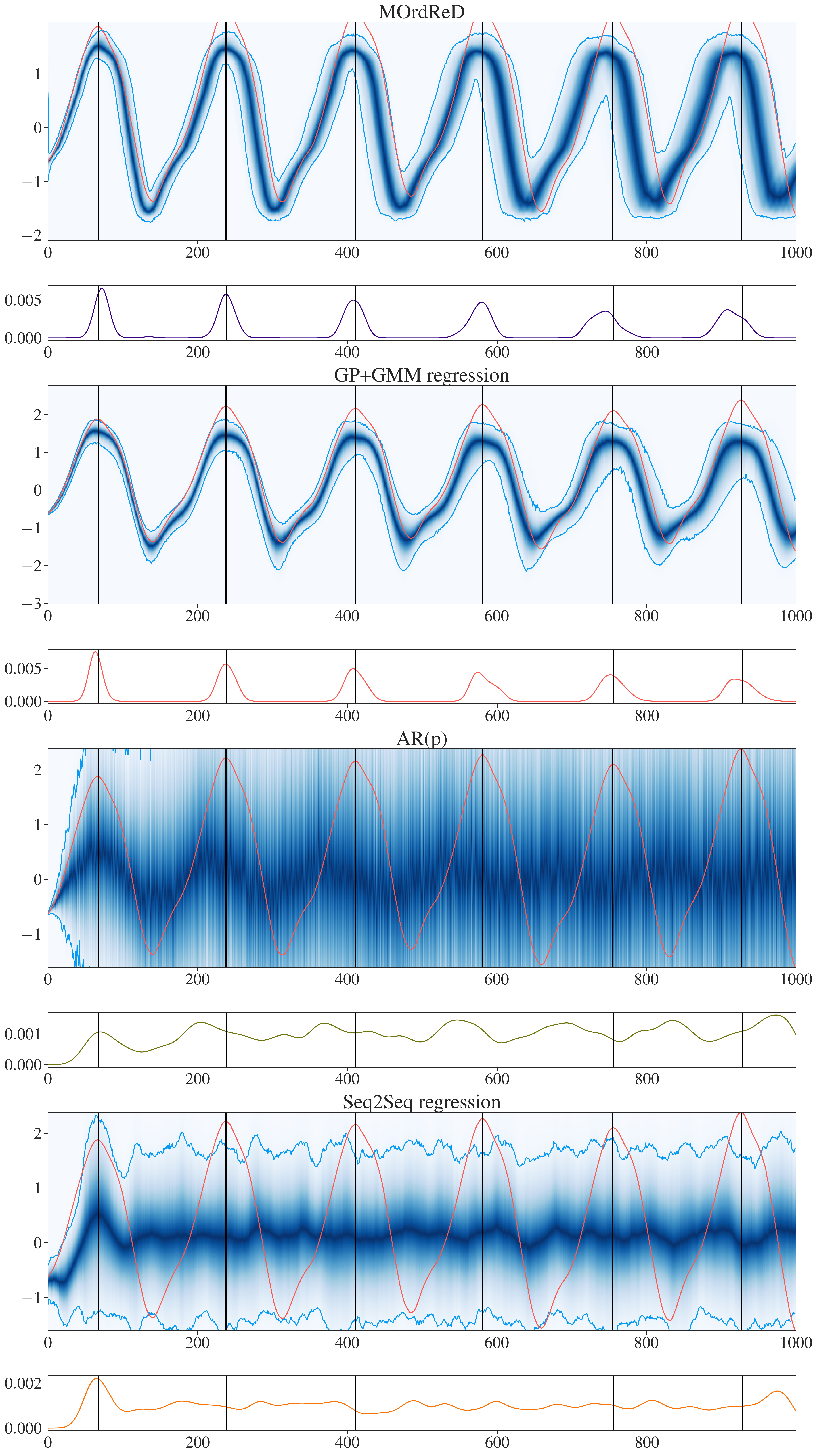}
	\caption{Example of out-of-sample forecast with predictive horizon $P_h=1000$ for the tide height time series dataset. This is an example of a dataset in which our model achieves performance comparable to the best-performing one, which is the GP baseline in this case. Ground truth (orange) overlayed on top of the predictive distributions given by each model. 95\% confidence bounds are highlighted in light blue. Maxima predictive densities are provided under each predictive distribution. Vertical lines were drawn at the optima of interest.}
	\label{fig:tide}
\end{figure*}

We further provide a visual example of these densities in Figures \ref{fig:heart} and \ref{fig:tide}. In the former, we observe that MOrdReD accurately predicts the timing of all maxima in the long-term, clearly outperforming all the other baselines in this task. On the contrary, in the case of the tide height dataset, we observe that the GP baseline is the one that accurately forecasts the timing of all maxima, since MOrdReD syncs out with respect to the ground truth towards the end of the forecast.

	\section{CONCLUDING REMARKS}
\label{sec:conclusions}

In this article we introduce a novel, fully end-to-end ordinal time series forecasting method based on recurrent neural networks. By recasting time series forecasting as an ordinal regression task, we demonstrate how state-of-the-art Machine Learning methods can be employed to perform reliable long-term time series prediction in a scalable and general-purpose fashion. We show how recent developments in the Bayesian neural networks literature can be used to quantify predictive uncertainty in time series forecasting, a major task that the literature of time series prediction with ANNs has paid relatively little attention to. Furthermore, our ordinal framework enables the description of adaptable multi-modal, non-Gaussian behaviour through time. 

Crucially, we highlight that these results are achieved in a fully end-to-end fashion. Little to no human intervention is needed to use MOrdReD for model design decisions. Direct learning from data is thence enabled without compromising scalability, assuming restrictive shapes of predictive distributions, or requiring expert craftsmanship for model design, as is often the case in designing a kernel for Gaussian Processes. 

In order to assess the performance of our framework, we provide a large-scale benchmark test over 45 different datasets drawn from an ample range of application domains and synthetic maps. Our 45 datasets were drawn from a variety of sources and include time series with quasi-periodic behaviour and complex shapes that in some cases are only modelled accurately by MOrdReD, as we exemplify with the electrocardiogram case. We compare our method with state-of-the-art baselines in the time series literature from both the Statistics and the Machine Learning perspectives, and evaluate a number of metrics to do with predictive accuracy, uncertainty quantification and forecast reliability.

We find that our model is empirically capable of outperforming other state-of-the-art competitors in terms of long-term forecasting uncertainty estimation, while also inheriting all the advantages of neural network models, such as scalability. As a result, our framework is consistently top-ranked in all metrics. We additionally provide evidence that it would also be the second-best performing model whenever another baseline achieved the best rank, and in a similar vein, we show that it is unlikely that it will be outperformed by all other baselines simultaneously in any given task, for any metric considered in the assessment.

Finally, we demonstrate how our model can be used to construct the predictive distribution of the occurrence of critical events of interest in applied settings. We show that our framework yields long-term, reliable confidence intervals even in safety-critical environments such as cardiology and meteorology.

This study, though comprehensive, still assumes the existence of enough data to fit our model, and only considers the case of bounded univariate time series autoregression. Future directions include extending this framework to the univariate case with exogenous observations, and more generally, to the multivariate case; as well as extensions that can achieve competitive performance even in presence of little data.
	\section{ACKNOWLEDGEMENTS}

The authors would like to thank Xiaowen Dong and other members of the Machine Learning Research Group at the University of Oxford for their feedback and edits on the paper clarity and structure.

\section{CONFLICT OF INTEREST}

The authors of this manuscript declare that there is no conflict of interest between this manuscript and other published works.

Bernardo P\'erez Orozco is funded by Consejo Nacional de Ciencia y Tecnolog\'ia (CONACYT) in Mexico CVU ID \#598304. Gabriele Abbati was supported by University of Oxford,
Google DeepMind. We also gratefully acknowledge the support provided by the Royal Academy of Engineering and the Oxford-Man Institute of Quantitative Finance.
	
	\printbibliography
	
	
	\newpage
\onecolumn
\section*{APPENDIX A}
In this appendix we provide a list of the 45 datasets used throughout the article. Those marked with an (*) were further used for the event occurrence forecasting task. Unless otherwise stated, these were all drawn from the compilation made by \cite{Fulcher20130048}.

\begin{enumerate}
    \item \{MACKEY\} Mackey-Glass chaotic attractor, synthesised by the authors according to the system:
	\begin{eqnarray}
	    x_{t+1} = (1 - b)x_t + a\frac{x_{t-\tau}}{1 + x_{t-\tau}^{n}} 
	\end{eqnarray}
	
	with $a=0.2, b=0.1, \tau=17, n=10$ and discarding the first 1,000 burnout samples.
	
	\item (*) \{ECG\} a recording of the electrical activity of the human heart \citep{rezek1998stochastic}, which highlights forecasting of a quasi-periodic, complex-shaped signal where accurate timing is essential;
	\item (*) \{AIRFLOW\} a recording of the breathing activity of a human\citep{rezek1998stochastic}, which highlights forecasting of a quasi-periodic signal where each quasi-cycle has a relatively long time period;
	\item (*) \{TIDE\} data taken from the Sotonmet \citep{Bramblemet} environmental sensors that record tide heights in an off-shore weather station. 
    \item \{AS\_s3.2\_\} birdsong excerpt from the Macaulay Animal Sounds Library,
    \item (*) \{CM\_air.s\} air temperature time series taken by different sensors in the Eurasia region between 1948 and 2007,
    \item \{CM\_air.2\} a second air temperature time series taken by different sensors in the Eurasia region between 1948 and 2007,
    \item \{CM\_prate\} precipitation rate time series taken by different sensors in the Eurasia region between 1948 and 2007,
    \item (*)\{CM\_slp19\} a sea level pressure time series taken by different sensors in the Eurasia region between 1948 and 2007,
    \item \{CM\_SLP.2\} a second sea level pressure time series taken by different sensors in the Eurasia region between 1948 and 2007,
    \item (*) \{CM\_rhum1\} relative humidity time series taken by different sensors in the Eurasia region between 1948 and 2007,
    \item \{CM\_lwtla\} Lamb/Jenkins weather type series measured from 1861 to 1997,
    \item \{EMexptqp\} Quasi-periodic output of Eric Weeks' Annulus experiment,
    \item \{EM\_henon\} Henon map 
    \begin{eqnarray}
        x' =& a + by - x^2 \\
        y' =& x
    \end{eqnarray}
    with $a=1.4, b=0.3$
    
    \item \{EMlorenz\} Lorenz map 
    \begin{eqnarray}
        x' =& \sigma (y-x)\\
        y' =& rx -y - xz\\ 
        z' =& xy - bz
    \end{eqnarray}
    with $\sigma=10, r=28, b=8/3$,
    
    \item \{EM\_rossl\} Rössler attractor
    \begin{eqnarray}
        x' =& -z-y \\ y' =& x + ay \\ z' =& b + z(x-c)
    \end{eqnarray}
    with $a=0.15, b=0.20, c=10.0$,
    
    \item \{FI\_yahoo\} Log returns of GPSC stock chart,
    \item \{FL\_ACT\_L\} $z-$channel of the ACT attractor:
    \begin{eqnarray}
        x' =& \alpha(x-y)\\ y' =& -4\alpha y + xz + \mu x^3 \\ z' =& -\delta\alpha z + xy + \beta z^2
    \end{eqnarray}
    computed with $\alpha=1.8, \beta=-0.07, \delta=1.5, \mu=0.02$,
    
    \item \{FL\_chen\_\} $x-$channel of Chen's system 
    \begin{eqnarray}
        x' =& a(y-x)\\ y' =& (c-a)x -xz + cy \\ z' =& xy - bz
    \end{eqnarray}
    computed with $a=35, b=3, c=28$,
    
    \item \{FL\_dblsc\} $y-$channel of the double-scroll system 
    \begin{eqnarray}
        x' =& y \\ y' =& z \\ z' =& -a[z + y + x - \text{sgn}(x)]
    \end{eqnarray}
    computed with $a=0.8$ and initial conditions $x_0=0.01, y_0=0.01, z_0=0$,
    
    \item \{FL\_hadle\} $x-$channel of the Hadley circulation system 
    \begin{eqnarray}
        x' =& -y^2 - z^2 -ax + aF\\ y' =&xy -bxz -y + G \\ z' =& bxy + xz - z
    \end{eqnarray}
    computed with $a=0.25, b=4, F=8, G=1$ and initial conditions $x_0=0, y_0=0, z_0=1.3$,
    
    \item \{FL\_labyr\} $y-$channel of the Labyrinth Chaos system 
    \begin{eqnarray}
        x' =& \sin(y) \\ y' =& -\sin(z) \\ z' =& \sin(x)
    \end{eqnarray}
    computed with initial conditions $x_0=0.1, y_0=0, z_0=0$,
    
    \item \{FL\_moore\} $z-$channel of the Moore-Spiegel oscillator 
    \begin{eqnarray}
        x' =& y \\ y' =& z \\ z' =& -z - (T-R + Rx^2)y - Tx
    \end{eqnarray}
    computed with $T=6, R=20$,
    
    \item \{FL\_noseh\} $z-$channel of the Nosé-Hoover oscillator 
    \begin{eqnarray}
        x' =& y \\ y' =& -x + yz \\ z' =& a - y^2
    \end{eqnarray}
    computed with $a=1$ and initial conditions $x_0=0, y_0=5, z_0=0$,
    
    \item \{FL\_ruckl\} $z-$channel of the Rucklidge attractor 
    \begin{eqnarray}
        x' =& -\kappa x + \lambda y - yz \\ y' =& x \\ z' =& -z + y^2
    \end{eqnarray}
    computed with $\kappa=2, \lambda=6.7$ and initial conditions $x_0=1, y_0=0, z_0=4.5$,
    
    \item \{FL\_simpq\} $y-$channel of the simplest quadratic flow 
    \begin{eqnarray}
        x' =& y \\ y' =& z \\ z' =& -az + y^2 - x
    \end{eqnarray}
    computed with $a=2.028$ and initial conditions $x_0=0.9, y_0=0, z_0=0.5$,
    
    \item \{FL\_thoma\} $y-$channel of Thomas cyclically symmetric attractor 
    \begin{eqnarray}
        x' =& -bx + \sin(y) \\ y' =& -by + \sin(z) \\ z'  -bz + \sin(x)
    \end{eqnarray}
    computed with $b=0.18$ and initial conditions $x_0=0.1, y_0=0, z_0=0$,
    
    \item \{FL\_windm\} $y-$channel of the WINDMI attractor 
    \begin{eqnarray}
        x' =& y \\ y' =& z \\ z' =& -az - y + b - \exp(x)
    \end{eqnarray}
    for $a=0.7, b=2.5$ and initial conditions  $x_0=0, y_0=0.8, z_0=0$,
    
    \item \{K\_standa\} Log returns of the Standard \& Poor index,
    \item \{MC\_inttr\} Internet traffic data from an ISP, provided by the Time Series Data Library,
    \item \{MP\_Lozi\_\} Lozi map $x_{t+1} = 1 - a|x_t| + bx_{n-1}$ with $a=1.7, b=0.5$ and initial conditions $x_1=-0.1, x_0=0.15$, 
    \item \{MP\_freit\} Freitas' stochastic sine map $x_{t+1}=\mu\sin{(x_t)} + Y_t\eta_t$, with parameters $\mu=2.4, b=3, q=0.2, \eta_t\sim\text{Uniform}(-b, b), Y_t\sim\text{Bernoulli}(q)$,
    \item \{MP\_logis\} Logistic map $x_{t+1} = Ax_t(1-x_t)$ with $A=3.2$ and initial condition $x_0=0.91$,
    \item \{MUS\_Si\_l\}  a music excerpt from B. Fulcher's personal collection, entitled \textit{Si loin de vous},
    \item \{MUS.3\_78\} a second music excerpt from B. Fulcher's personal collection, entitled \textit{3},
    \item \{SFX\_mach\} Sound Jay Mach Electric Drill sound effect,
    \item \{SF\_Acont\}  Santa Fe laser generated data excerpt,
    \item \{SF\_B1\_1\} Santa Fe heart rate data excerpt,
    \item \{SF\_D1\}  Santa Fe synthetically generated snippet,
    \item \{SL\_perci\} StatLib demeaned water level measurements,
    \item \{SPIDR\_hp\} Hemispheric Power Index excerpt provided by the Space Physics Interactive Data Resource,
    \item \{SY\_AR2\_T\} Timmer nonstationary autoregressive process with $\tau=20, T_{mean}=20, T_{mod}=20, \sigma=1, \mathcal{M}_{T}=5, \eta=1000$,
    \item \{SY\_NLAR2\} Faes nonlinear autoregressive process with $a_1=3.6, a_2=0.8$,
    \item \{TSAR\_eqe\} Earthquake and explosion seismic series provided in Stoffer's \textit{Time Series Analysis and its applications, with R examples}.
    \item \{TXT\_slc\_\} Project Gutenberg excerpt of Dickens' \textit{Oliver Twist}.
\end{enumerate}

	\onecolumn
\section*{APPENDIX B}
In this appendix we provide the full results for the performance of each model on each dataset for all metrics. Full details of the benchmarking task and the definition of our metrics are given in Section \ref{sec:experiments}.

\begin{figure*}[!ht]
		\centering
		\includegraphics[width=5in]{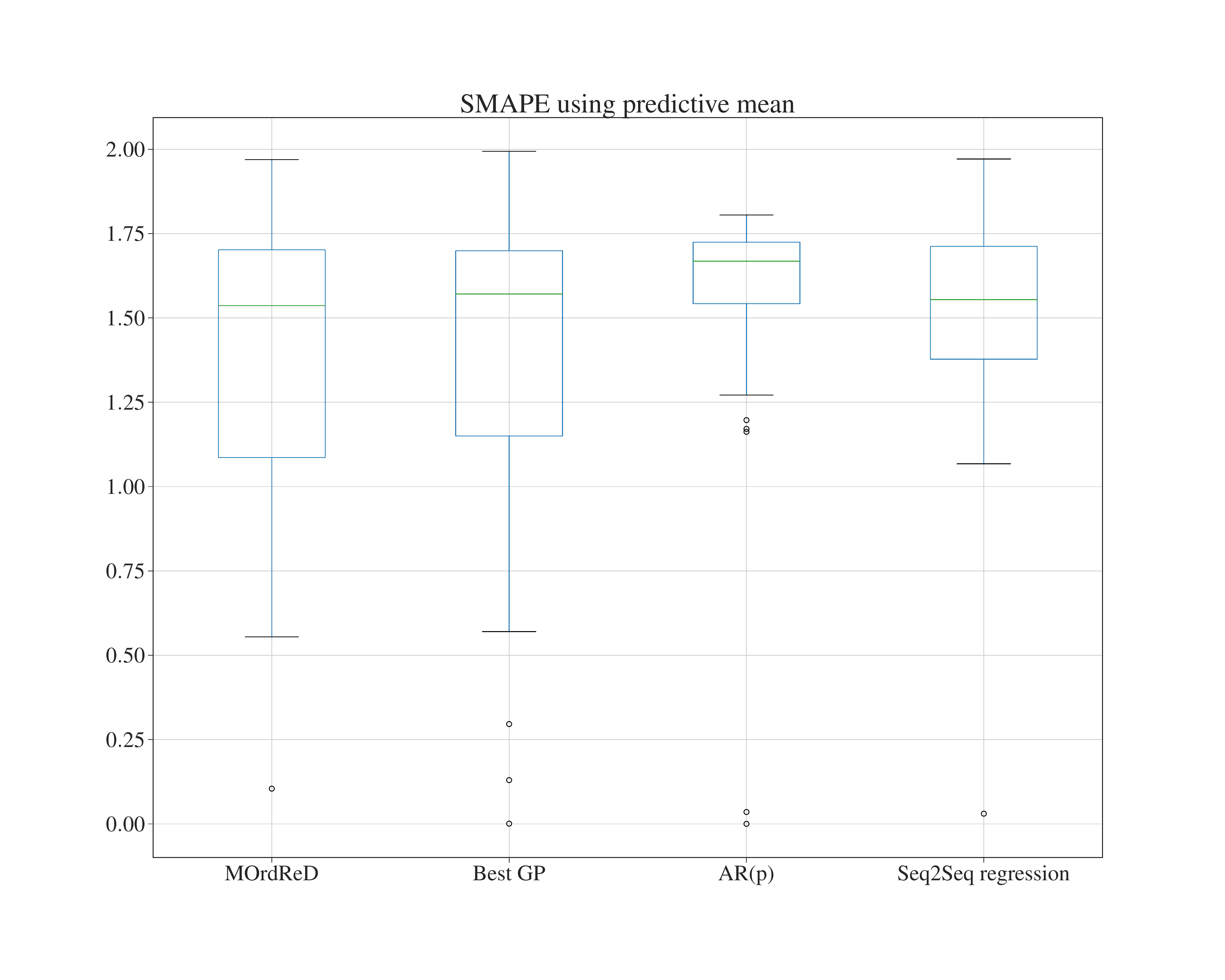}
	\caption{Box plot for the SMAPE deviation of the predictive distribution's mean. We can see that GPs and MOrdReD achieve a similar performance, clearly surpassing the other two baselines.}
	\label{fig:box_smape_mean}
\end{figure*}
~
\begin{figure*}
		\centering
		\includegraphics[width=5in]{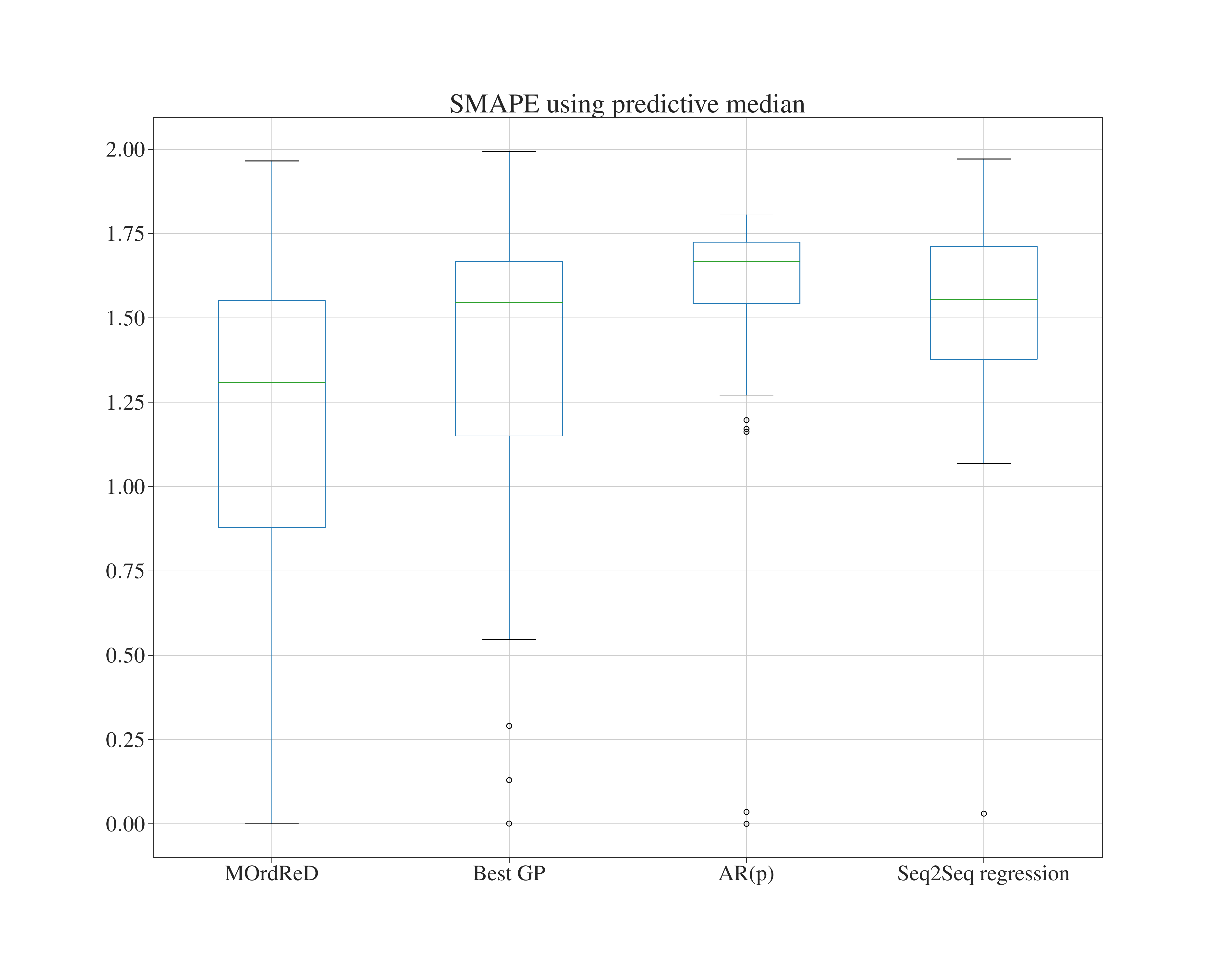}
	\caption{Box plot for the SMAPE deviation of the predictive distribution's median. We can see that GPs and MOrdReD achieve a similar performance, clearly surpassing the other two baselines.}
	\label{fig:box_smape_median}
\end{figure*}
~
\begin{figure*}
		\centering
		\includegraphics[width=5in]{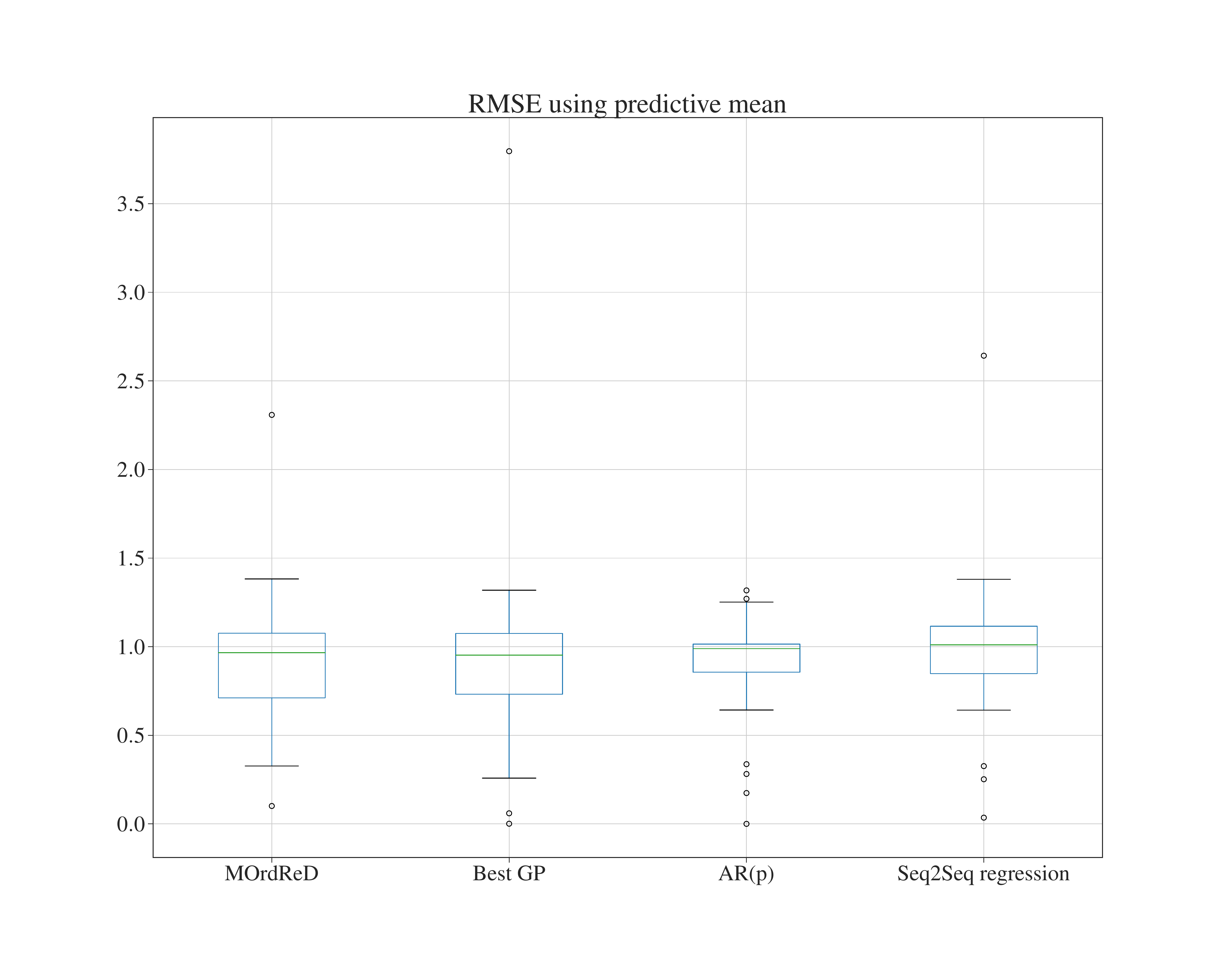}
	\caption{Box plot for the RMSE deviation of the predictive distribution's mean. We can see that GPs and MOrdReD achieve a similar performance, clearly surpassing the other two baselines.}
	\label{fig:box_rmse_mean}
\end{figure*}
~
\begin{figure*}
		\centering
		\includegraphics[width=5in]{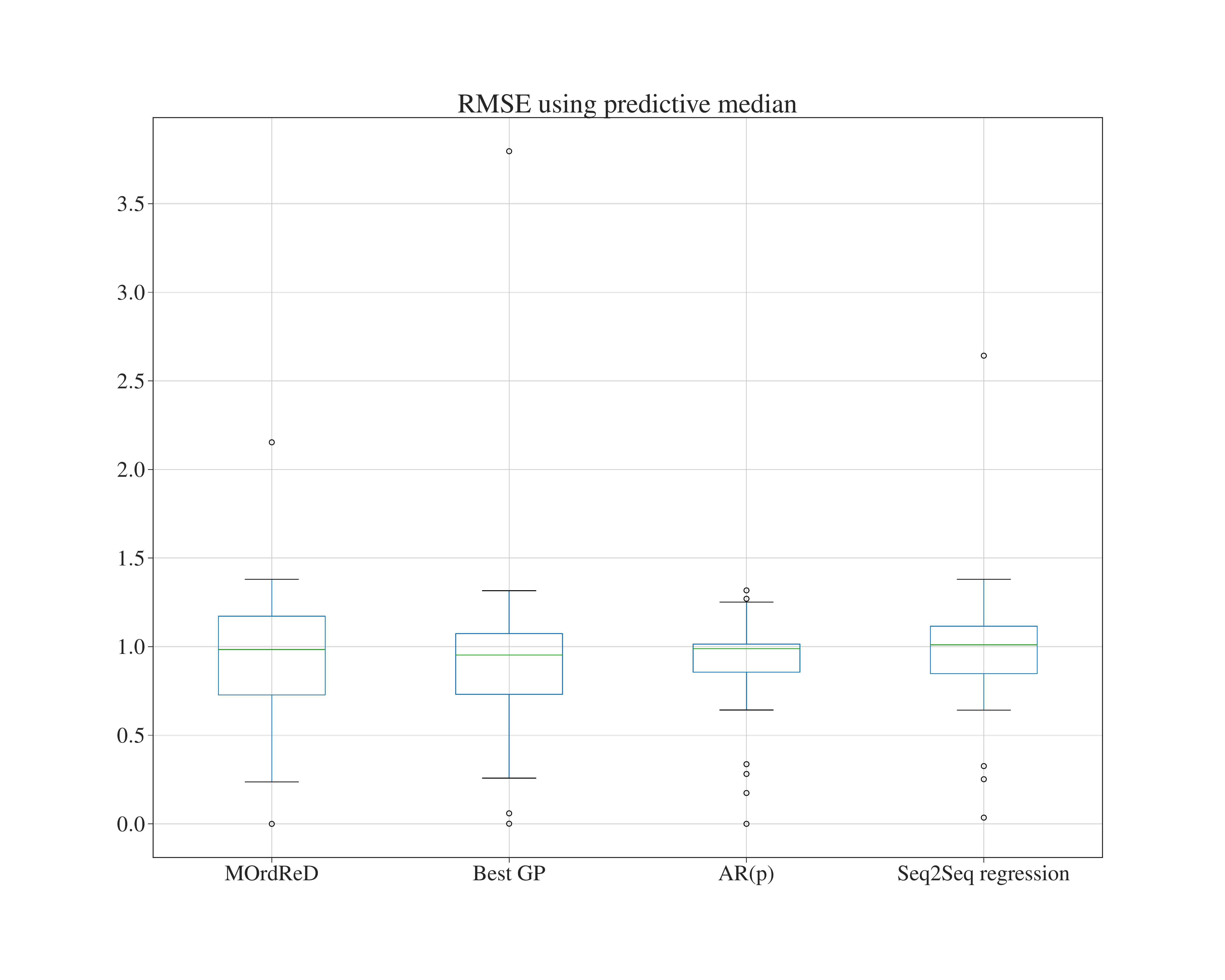}
	\caption{Box plot for the RMSE deviation of the predictive distribution's median. We can see that GPs and MOrdReD achieve a similar performance, clearly surpassing the other two baselines.}
	\label{fig:box_rmse_median}
\end{figure*}
~
\begin{figure*}
		\centering
		\includegraphics[width=5in]{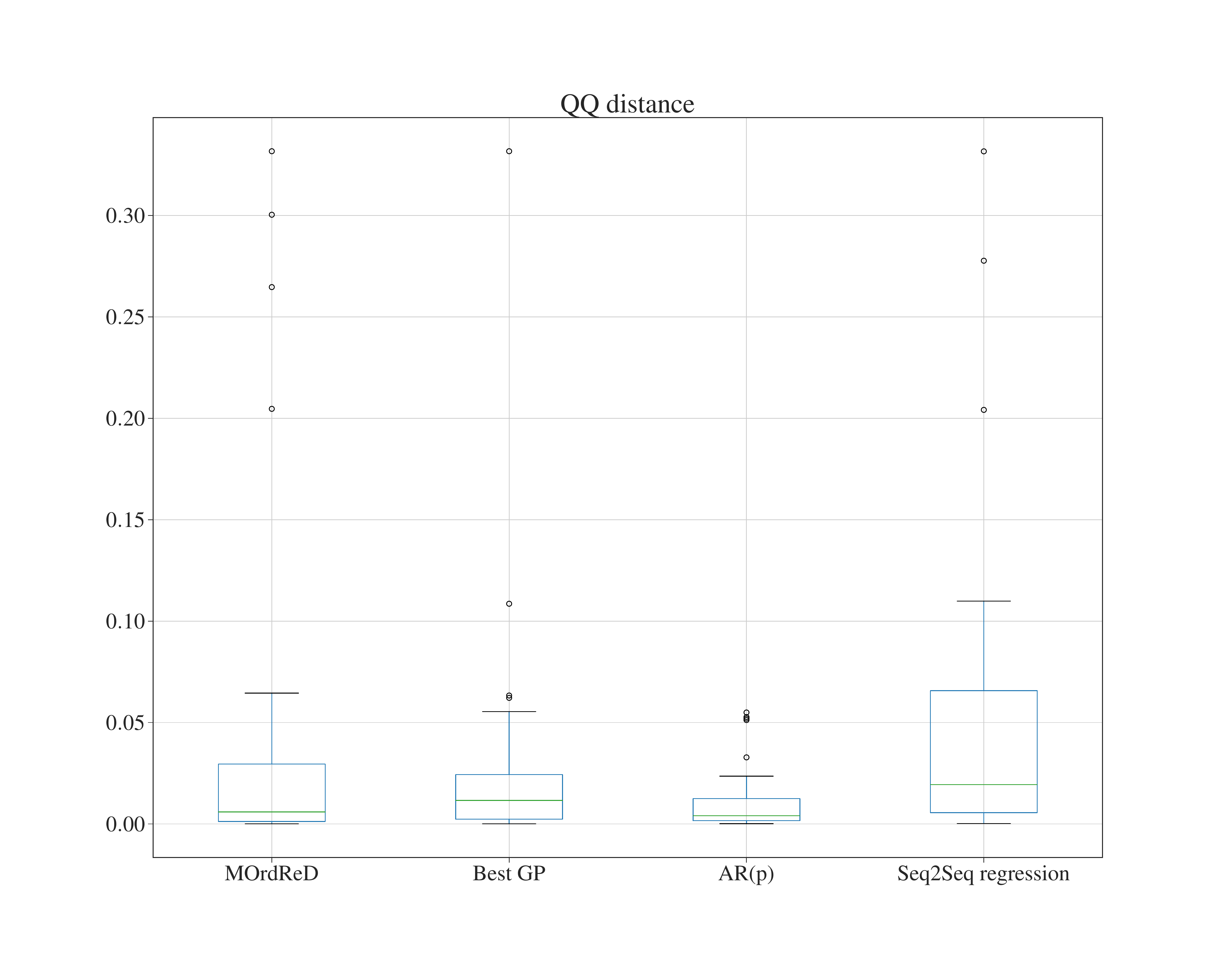}
	\caption{Box plot for the QQ distance results for the full prediction up to time index $P_h=1000$.}
	\label{fig:box_qq}
\end{figure*}
~
\begin{figure*}
		\centering
		\includegraphics[width=5in]{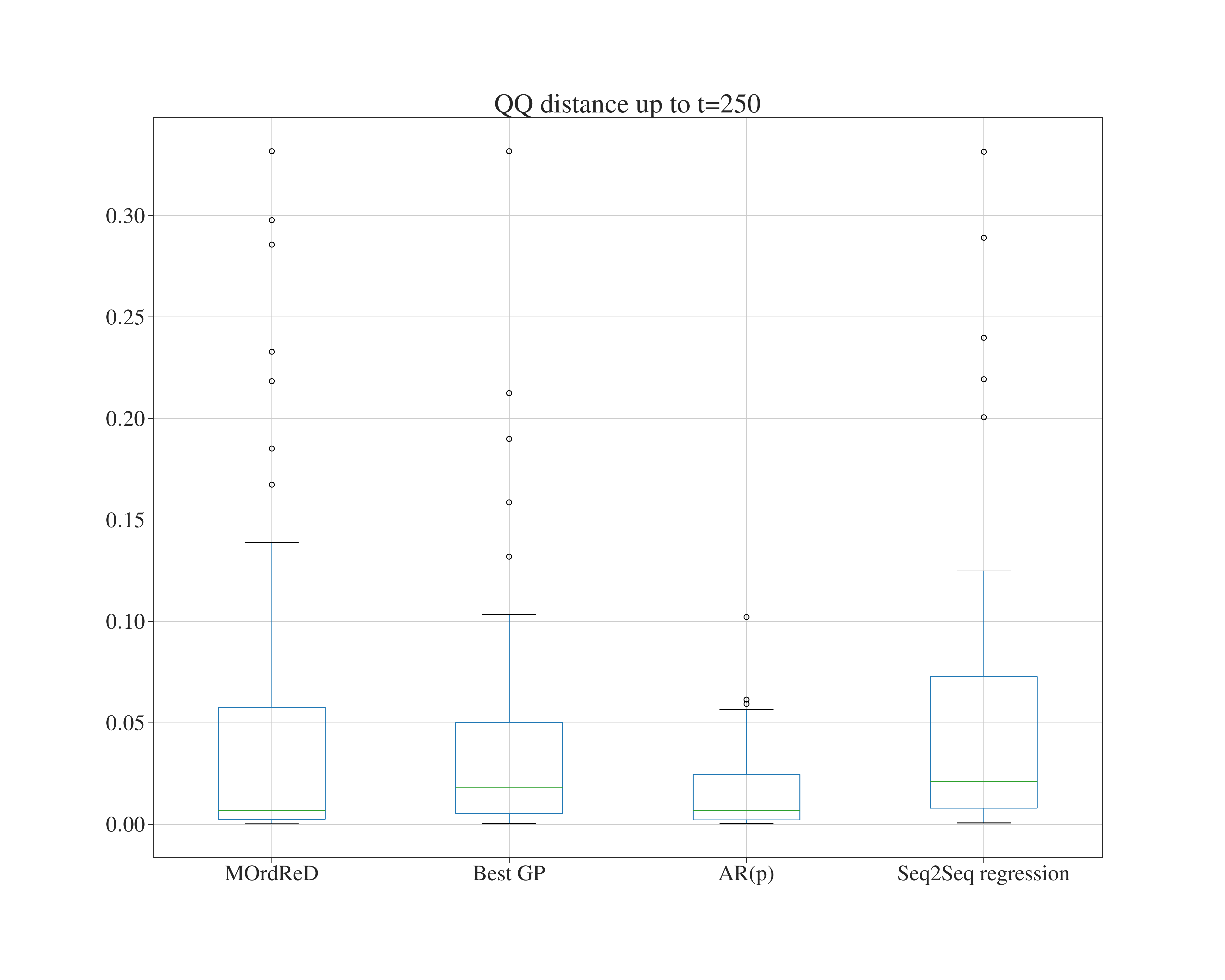}
	\caption{Box plot for the QQ distance results up to time index 250.}
	\label{fig:box_qq_250}
\end{figure*}
~
\begin{figure*}[!h]
		\centering
		\includegraphics[width=5.75in]{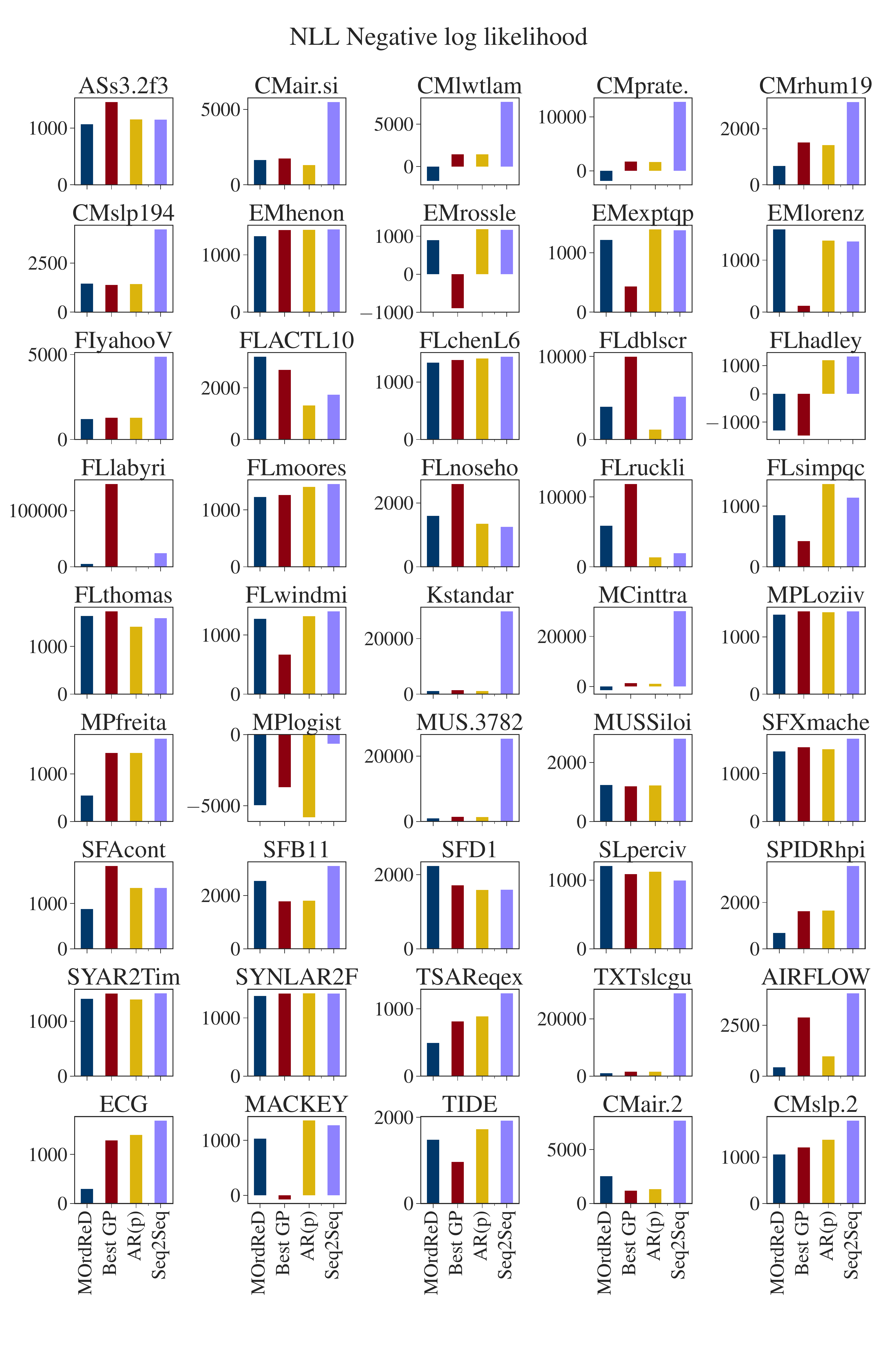}
	\caption{Bar chart for the NLL results for each model and each dataset.}
	\label{fig:bar_nll}
\end{figure*}
~
\begin{figure*}[!h]
		\centering
		\includegraphics[width=5.75in]{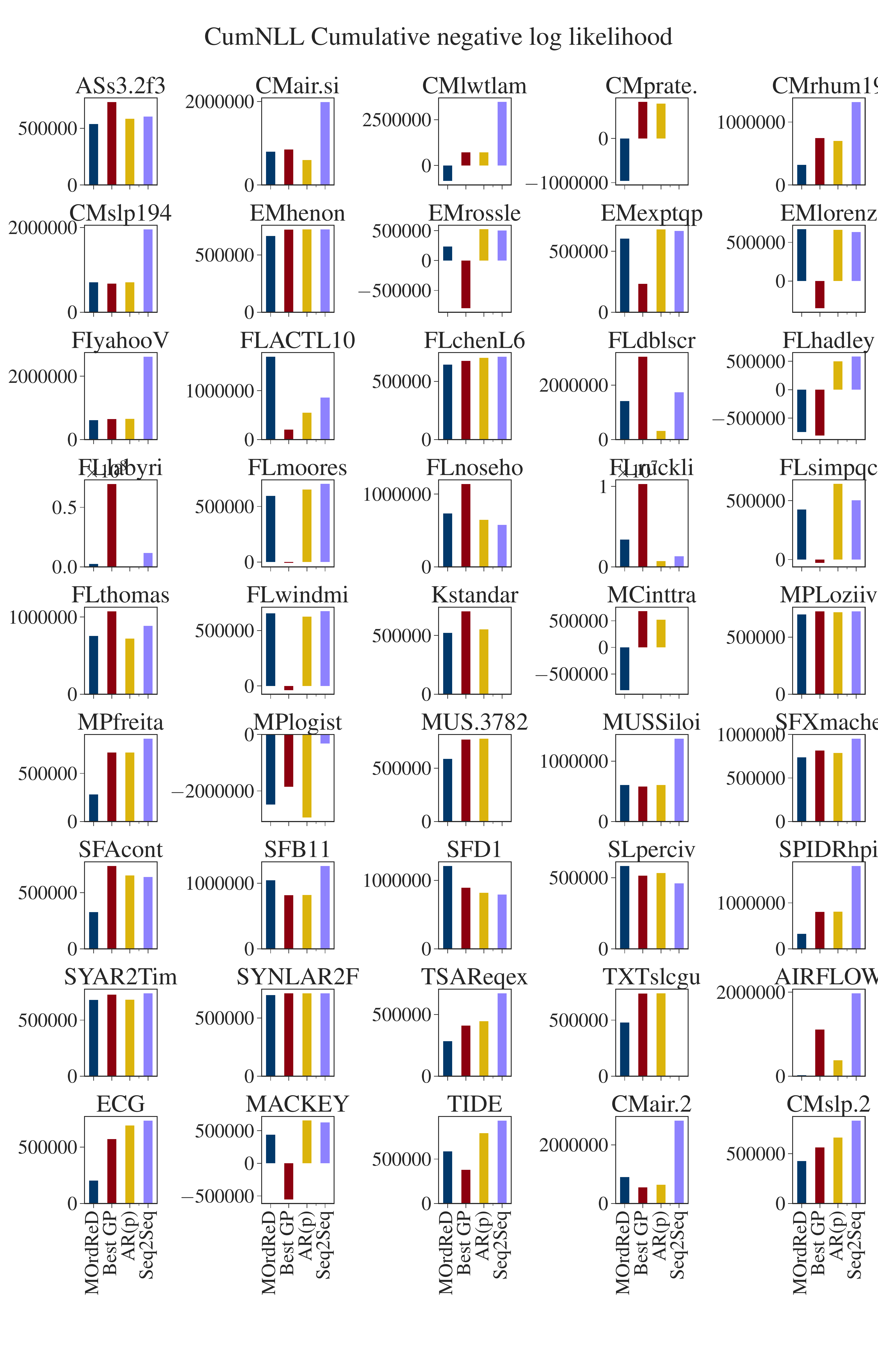}
	\caption{Bar chart for the Cumulative NLL results for each model and each dataset.}
	\label{fig:bar_cumnll}
\end{figure*}

	\clearpage
\onecolumn
\section*{APPENDIX C}
In this appendix we provide the optimised hyperparameters for our MOrdReD and AR(p) models. 

\begin{table*}[!ht]
\centering
\begin{tabular}{lrlr}
\toprule
{} &   Order & & Order\\
\midrule
\textbf{AS\_s3.2\_} &  16 & \textbf{MP\_freit} &  16\\
\textbf{CM\_air.s} &  64 & \textbf{MP\_logis} &  64\\
\textbf{CM\_lwtla} &  16 & \textbf{MUS.3\_78} &  16 \\
\textbf{CM\_prate} &  64 & \textbf{MUS.3\_78} &  16 \\
\textbf{CM\_rhum1} &  64 & \textbf{MUS.3\_78} &  16  \\
\textbf{CM\_slp19} &  64 & \textbf{SFX\_mach} &  16 \\
\textbf{EM\_henon} &  16 & \textbf{SF\_Acont} &  64 \\
\textbf{EM\_rossl} &  16 & \textbf{SF\_B1\_1 } &  64 \\
\textbf{EMexptqp} &  16 & \textbf{SF\_D1   } &  16 \\
\textbf{EMlorenz} &  16 & \textbf{SL\_perci} &  64\\
\textbf{FI\_yahoo} &  64 & \textbf{SPIDR\_hp} &  64 \\
\textbf{FL\_ACT\_L} &  16 & \textbf{SY\_AR2\_T} &  64\\
\textbf{FL\_chen\_} &  16 & \textbf{SY\_NLAR2} &  64\\
\textbf{FL\_dblsc} &  16 & \textbf{TSAR\_eqe} &  16\\
\textbf{FL\_hadle} &  16 & \textbf{TXT\_slc\_} &  64\\
\textbf{FL\_labyr} &  16 & \textbf{AIR     } &  16\\
\textbf{FL\_moore} &  16 & \textbf{ECG   } &  32\\
\textbf{FL\_noseh} &  16 & \textbf{MACKEY      } &  16 \\
\textbf{FL\_ruckl} &  16 & \textbf{TIDE    } &  16 \\
\textbf{FL\_simpq} &  16 & \textbf{CM\_air 2} &  64\\
\textbf{FL\_thoma} &  16 & \textbf{CM\_slp 2} &  64\\
\textbf{FL\_windm} &  16 &  & \\
\textbf{K\_standa} &  64 & & \\
\textbf{MC\_inttr} &  64 & & \\
\textbf{MP\_Lozi\_} &  16 & & \\

\bottomrule
\end{tabular}
\caption{Best value for the $p$ hyperparameter of each AR(p) model found by grid search on $p\in\{16,32,64\}$}
\label{tab:ar_hyper}
\end{table*}

\begin{table*}
\centering
\begin{tabular}{lrrrr}
\toprule
{} &    Hidden Units  $h_u$ &  Dropout rate $p_{\text{dropout}}$&    L2 regularisation $\lambda$ &  Ordinal bins \\
\midrule
\textbf{AS\_s3.2\_} &   64.0 &         0.25 &  1e-08 &        300.0 \\
\textbf{CM\_air.s} &   64.0 &         0.25 &  1e-08 &        300.0 \\
\textbf{CM\_lwtla} &  128.0 &         0.25 &  1e-07 &        300.0 \\
\textbf{CM\_prate} &  128.0 &         0.25 &  1e-07 &        300.0 \\
\textbf{CM\_rhum1} &  128.0 &          0.5 &  1e-07 &        300.0 \\
\textbf{CM\_slp19} &  256.0 &          0.5 &  1e-06 &        300.0 \\
\textbf{EM\_henon} &   64.0 &          0.5 &  1e-06 &        300.0 \\
\textbf{EM\_rossl} &  320.0 &         0.25 &  1e-07 &        300.0 \\
\textbf{EMexptqp} &  256.0 &          0.5 &  1e-08 &        300.0 \\
\textbf{EMlorenz} &   64.0 &          0.5 &  1e-06 &        300.0 \\
\textbf{FI\_yahoo} &  320.0 &          0.5 &  1e-08 &        300.0 \\
\textbf{FL\_ACT\_L} &  256.0 &          0.5 &  1e-07 &        300.0 \\
\textbf{FL\_chen\_} &  128.0 &          0.5 &  1e-07 &        300.0 \\
\textbf{FL\_dblsc} &  128.0 &         0.25 &  1e-08 &        300.0 \\
\textbf{FL\_hadle} &  320.0 &         0.25 &  1e-07 &        300.0 \\
\textbf{FL\_labyr} &   64.0 &         0.25 &  1e-08 &        300.0 \\
\textbf{FL\_moore} &  320.0 &          0.5 &  1e-06 &        300.0 \\
\textbf{FL\_noseh} &  128.0 &         0.25 &  1e-08 &        300.0 \\
\textbf{FL\_ruckl} &  320.0 &          0.5 &  1e-07 &        300.0 \\
\textbf{FL\_simpq} &   64.0 &          0.5 &  1e-08 &        300.0 \\
\textbf{FL\_thoma} &   64.0 &          0.5 &  1e-06 &        300.0 \\
\textbf{FL\_windm} &  256.0 &          0.5 &  1e-08 &        300.0 \\
\textbf{K\_standa} &  256.0 &          0.5 &  1e-06 &        300.0 \\
\textbf{MC\_inttr} &  128.0 &         0.25 &  1e-06 &        300.0 \\
\textbf{MP\_Lozi\_} &   64.0 &          0.5 &  1e-08 &        300.0 \\
\textbf{MP\_freit} &   64.0 &         0.25 &  1e-07 &        300.0 \\
\textbf{MP\_logis} &  128.0 &         0.25 &  1e-06 &        300.0 \\
\textbf{MUS.3\_78} &  320.0 &          0.5 &  1e-07 &        300.0 \\
\textbf{MUS\_Si\_l} &  320.0 &          0.5 &  1e-08 &        300.0 \\
\textbf{SFX\_mach} &   64.0 &          0.5 &  1e-08 &        300.0 \\
\textbf{SF\_Acont} &   64.0 &         0.25 &  1e-06 &        236.0 \\
\textbf{SF\_B1\_1 } &  128.0 &          0.5 &  1e-08 &        300.0 \\
\textbf{SF\_D1   } &  320.0 &          0.5 &  1e-08 &        300.0 \\
\textbf{SL\_perci} &  256.0 &          0.5 &  1e-06 &        300.0 \\
\textbf{SPIDR\_hp} &  128.0 &          0.5 &  1e-08 &        300.0 \\
\textbf{SY\_AR2\_T} &   64.0 &         0.25 &  1e-06 &        300.0 \\
\textbf{SY\_NLAR2} &  320.0 &          0.5 &  1e-06 &        300.0 \\
\textbf{TSAR\_eqe} &  128.0 &         0.25 &  1e-07 &        300.0 \\
\textbf{TXT\_slc\_} &  256.0 &         0.25 &  1e-07 &        300.0 \\
\textbf{AIRFLOW} &  256.0 &          0.5 &  1e-06 &        226.0 \\
\textbf{ECG   } &  256.0 &          0.5 &  1e-07 &        134.0 \\
\textbf{MACKEY      } &  256.0 &         0.25 &  1e-07 &        300.0 \\
\textbf{TIDE    } &  256.0 &          0.5 &  1e-06 &        300.0 \\
\textbf{CM\_air.2} &  128.0 &         0.25 &  1e-07 &        300.0 \\
\textbf{CM\_SLP.2} &  320.0 &         0.25 &  1e-07 &        259.0 \\
\bottomrule
\end{tabular}
\caption{Best value for the hyperparameters of our MOrdReD models found by grid search on $h_u\in\{64,128,256,320\},\lambda\in\{1e-6,1e-7,1e-8\},p_{\text{dropout}}\in\{0.25,0.35,0.5\}$.}
\label{tab:mordred_hyper}
\end{table*}

\begin{table*}
\centering
\begin{tabular}{lrrr}
\toprule
{} &    Hidden Units  $h_u$ &  Dropout rate $p_{\text{dropout}}$&    L2 regularisation $\lambda$ \\
\midrule
\textbf{AS\_s3.2\_} &  128.0 &          0.5 &  1e-08 \\
\textbf{CM\_air.s} &  320.0 &          0.5 &  1e-07 \\
\textbf{CM\_lwtla} &  256.0 &          0.5 &  1e-08 \\
\textbf{CM\_prate} &   64.0 &         0.25 &  1e-06 \\
\textbf{CM\_rhum1} &  128.0 &          0.5 &  1e-07 \\
\textbf{CM\_slp19} &  128.0 &          0.5 &  1e-06 \\
\textbf{EM\_henon} &   64.0 &          0.5 &  1e-06 \\
\textbf{EM\_rossl} &  128.0 &         0.25 &  1e-07 \\
\textbf{EMexptqp} &  256.0 &         0.25 &  1e-07 \\
\textbf{EMlorenz} &  128.0 &         0.25 &  1e-07 \\
\textbf{FI\_yahoo} &  128.0 &         0.25 &  1e-08 \\
\textbf{FL\_ACT\_L} &  128.0 &         0.25 &  1e-06 \\
\textbf{FL\_chen\_} &  320.0 &          0.5 &  1e-08 \\
\textbf{FL\_dblsc} &  256.0 &         0.25 &  1e-08 \\
\textbf{FL\_hadle} &  128.0 &         0.25 &  1e-06 \\
\textbf{FL\_labyr} &  320.0 &         0.25 &  1e-07 \\
\textbf{FL\_moore} &  128.0 &         0.25 &  1e-07 \\
\textbf{FL\_noseh} &  256.0 &          0.5 &  1e-08 \\
\textbf{FL\_ruckl} &  256.0 &         0.25 &  1e-06 \\
\textbf{FL\_simpq} &  128.0 &         0.25 &  1e-08 \\
\textbf{FL\_thoma} &  320.0 &         0.25 &  1e-07 \\
\textbf{FL\_windm} &  256.0 &          0.5 &  1e-08 \\
\textbf{K\_standa} &   64.0 &         0.25 &  1e-06 \\
\textbf{MC\_inttr} &   64.0 &         0.25 &  1e-06 \\
\textbf{MP\_Lozi\_} &   64.0 &         0.25 &  1e-08 \\
\textbf{MP\_freit} &   64.0 &         0.25 &  1e-06 \\
\textbf{MP\_logis} &   64.0 &         0.25 &  1e-07 \\
\textbf{MUS.3\_78} &   64.0 &         0.25 &  1e-06 \\
\textbf{MUS\_Si\_l} &  320.0 &          0.5 &  1e-08 \\
\textbf{SFX\_mach} &  320.0 &         0.25 &  1e-08 \\
\textbf{SF\_Acont} &  128.0 &         0.25 &  1e-07 \\
\textbf{SF\_B1\_1 } &  320.0 &         0.25 &  1e-06 \\
\textbf{SF\_D1   } &  256.0 &          0.5 &  1e-07 \\
\textbf{SL\_perci} &  128.0 &         0.25 &  1e-06 \\
\textbf{SPIDR\_hp} &  320.0 &         0.25 &  1e-08 \\
\textbf{SY\_AR2\_T} &  320.0 &         0.25 &  1e-08 \\
\textbf{SY\_NLAR2} &  128.0 &         0.25 &  1e-07 \\
\textbf{TSAR\_eqe} &  128.0 &          0.5 &  1e-06 \\
\textbf{TXT\_slc\_} &   64.0 &         0.25 &  1e-06 \\
\textbf{AIRFLOW     } &   64.0 &         0.25 &  1e-06 \\
\textbf{ECG   } &  128.0 &         0.25 &  1e-07 \\
\textbf{MACKEY      } &  128.0 &         0.25 &  1e-07 \\
\textbf{TIDE    } &  320.0 &         0.25 &  1e-07 \\
\textbf{CM\_air.2} &  256.0 &          0.5 &  1e-06 \\
\textbf{CM\_SLP.2} &  320.0 &          0.5 &  1e-06 \\
\bottomrule
\end{tabular}
\caption{Best value for the hyperparameters of our direct regression neural-network models found by grid search on $h_u\in\{64,128,256,320\},\lambda\in\{1e-6,1e-7,1e-8\},p_{\text{dropout}}\in\{0.25,0.35,0.5\}$.}
\label{tab:seq2seq_hyper}
\end{table*}

\end{document}